\documentclass[10pt,twocolumn,letterpaper]{article}

\usepackage{cvpr}              % To produce the CAMERA-READY version
\usepackage[linesnumbered,ruled,vlined]{algorithm2e}
\usepackage{graphicx}
\usepackage{amsmath}
\usepackage{amssymb}
\usepackage{booktabs}
\usepackage{amssymb}
\usepackage{amsmath,amsfonts}
\usepackage{amsopn}
\usepackage{bm} % bold symbol
\usepackage{multirow}
\usepackage{flushend}
\usepackage{tabularx}

\newcommand{\HT}{\method{Head2Toe}\xspace}

\newcommand{\vct}[1]{\boldsymbol{#1}} % vector
\newcommand{\mat}[1]{\boldsymbol{#1}} % matrix
\newcommand{\cst}[1]{\mathsf{#1}}  % constant

\newcommand{\field}[1]{\mathbb{#1}}
\newcommand{\R}{\field{R}} % real domain
\newcommand{\ProbOpr}[1]{\mathbb{#1}}
\newcommand{\expect}[2]{%
\ifthenelse{\equal{#2}{}}{\ProbOpr{E}_{#1}}
{\ifthenelse{\equal{#1}{}}{\ProbOpr{E}\left[#2\right]}{\ProbOpr{E}_{#1}\left[#2\right]}}} % Expectation: syntax: E{1}{2} = E_1[2], E{}{2}=E[2], E{1}{} = E_1
\newcommand{\vp}{\vct{p}}

\newcommand{\vx}{{\vct{x}}}

\newcommand{\mW}{\mat{W}}
\newcommand{\mV}{\mat{V}}
\newcommand{\mQ}{\mat{Q}}

\newcommand{\mS}{\mat{S}}

\newcommand{\mZ}{\mat{Z}}
\newcommand{\mP}{\mat{P}}

\newcommand{\mI}{\mat{I}}
\newcommand{\mK}{\mat{K}}

\newcommand{\sD}{\mathcal{D}}
\newcommand{\sU}{\mathcal{U}}

\newcommand{\eat}[1]{}
\newcommand{\method}[1]{\textsc{#1}}

\usepackage{enumitem}
\usepackage{arydshln}
\usepackage{subcaption} 
\usepackage[dvipsnames]{xcolor}
\usepackage{makecell}

\usepackage[pagebackref,breaklinks,colorlinks]{hyperref}

\newcommand{\zheda}[1]{{\color{magenta}[Zheda: #1]}}
\newcommand{\WLC}[1]{{\color{cyan} Harry: #1}\xspace}

\usepackage[capitalize]{cleveref}
\crefname{section}{Sec.}{Secs.}
\Crefname{section}{Section}{Sections}
\Crefname{table}{Table}{Tables}
\crefname{table}{Tab.}{Tabs.}
\newcommand\mypara[1]{\vspace{0.6mm}\noindent\textbf{#1}}

\def\cvprPaperID{10215} % *** Enter the CVPR Paper ID here
\def\confName{CVPR}
\def\confYear{2023}

\begin{document}

% \title{\LaTeX\ Author Guidelines for \confName~Proceedings}
\title{Visual Query Tuning: Towards Effective Usage of Intermediate Representations for Parameter and Memory Efficient Transfer Learning}
% for Parameter and Memory Efficient Transfer Learning

%\title{Visual Query Tuning: Decoupling the Role of Prompts in Feature Adaptation\\ and Feature Combination for Better Transfer Learning}

%\title{Summary Prompt Tuning: Decoupling the Role of Prompt in Feature Adaptation and Combination in Visual Recognition}

\iffalse
\author{
Cheng-Hao Tu\thanks{Equal contributions.}\\
The Ohio State University\\
\and
Zheda Mai\footnotemark[1]\\
The Ohio State University\\
\and
Wei-Lun Chao\\
The Ohio State University\\
}
\fi

\author{
Cheng-Hao Tu\thanks{Equal contributions.}
\and
Zheda Mai\footnotemark[1]
\and
Wei-Lun Chao
\and 
The Ohio State University, $\quad\{$tu.343, mai.145, chao.209$\}$@osu.edu
}

\maketitle

%%%%%%%%% ABSTRACT
\begin{abstract}
Intermediate features of a pre-trained model have been shown informative for making accurate predictions on downstream tasks, even if the model backbone is kept frozen. The key challenge is how to utilize these intermediate features given their gigantic amount. We propose visual query tuning (VQT), a simple yet effective approach to aggregate intermediate features of Vision Transformers. Through introducing a handful of learnable ``query'' tokens to each layer, VQT leverages the inner workings of Transformers to ``summarize'' rich intermediate features of each layer, which can then be used to train the prediction heads of downstream tasks. As VQT keeps the intermediate features intact and only learns to combine them, it enjoys memory efficiency in training, compared to many other parameter-efficient fine-tuning approaches that learn to adapt features and need back-propagation through the entire backbone. This also suggests the complementary role between VQT and those approaches in transfer learning. Empirically, VQT consistently surpasses the state-of-the-art approach that utilizes intermediate features for transfer learning and outperforms full fine-tuning in many cases. Compared to parameter-efficient approaches that adapt features, VQT achieves much higher accuracy under memory constraints. Most importantly, VQT is compatible with these approaches to attain even higher accuracy, making it a simple add-on to further boost transfer learning. Code is available at \url{https://github.com/andytu28/VQT}.
\end{abstract}

\section{Introduction}
\label{sec:intro}

%\subsection{Vervion 2}
  
Transfer learning by adapting large pre-trained models to downstream tasks has been a de facto standard for competitive performance, especially when downstream tasks have limited data \cite{zhuang2020comprehensive, lu2020knowledge}. Generally speaking, there are two ways to adapt a pre-trained model~\cite{kornblith2019better, evci2022head2toe}: updating the model backbone for new feature embeddings (the output of the penultimate layer) or recombining the existing feature embeddings, which correspond to the two prevalent approaches, \textit{fine-tuning} and \textit{linear probing}, respectively. \textit{Fine-tuning}, or more specifically, \textit{full fine-tuning}, updates all the model parameters end-to-end based on the new dataset. Although \textit{fine-tuning} consistently outperforms \textit{linear probing} on various tasks \cite{zhai2019large}, it requires running gradient descent for all parameters and storing a separate fine-tuned model for each task, making it computationally expensive and parameter inefficient. These problems become more salient with Transformer-based models whose parameters grow exponentially \cite{khan2022transformers,han2022survey,vaswani2017attention}. Alternatively, \textit{linear probing} only trains and stores new prediction heads to recombine features while keeping the backbone frozen. Despite its computational and parameter efficiency, \textit{linear probing} is often less attractive due to its inferior performance.

Several recent works have attempted to overcome such a dilemma in transfer learning. One representative work is by Evci~\etal \cite{evci2022head2toe}, who attributed the success of \textit{fine-tuning} to leveraging the ``intermediate'' features of pre-trained models and proposed to directly allow \textit{linear probing} to access the intermediate features. Some other works also demonstrated the effectiveness of such an approach~\cite{eom2022layover,evci2022head2toe}. Nevertheless, given numerous intermediate features in each layer, most of these methods require pooling to reduce the dimensionality, which likely would  eliminate useful information before the prediction head can access it.
%\zheda{"work" should be in plural?}
 
To better utilize intermediate features, we propose \textbf{Visual Query Tuning (VQT)}, a simple yet effective approach to aggregate the intermediate features of Transformer-based models like Vision Transformers (ViT) \cite{dosovitskiy2021an}. A Transformer usually contains multiple Transformer layers, each starting with a Multi-head self-attention (MSA) module operating over the intermediate feature tokens (often $>100$ tokens) outputted by the previous layer. The MSA module transforms each feature token by querying all the other tokens, followed by a weighted combination of their features. 
 
Taking such inner workings into account, \textbf{VQT} introduces a handful of \emph{learnable} ``query'' tokens to each layer, which, through the MSA module, can then ``summarize'' the intermediate features of the previous layer to reduce the dimensionality.
The output features of these query tokens after each layer can then be used by \emph{linear probing} to make predictions. Compared to pooling which simply averages the features over tokens, \textbf{VQT} performs a weighted combination whose weights are adaptive, conditioned on the features and the learned query tokens, and is more likely to capture useful information for the downstream task.

At first glance, \textbf{VQT} may look superficially similar to Visual Prompt Tuning (VPT)~\cite{jia2022vpt}, a recent transfer learning method that also introduces additional learnable tokens (\ie, prompts) to each layer of Transformers, but they are fundamentally different in two aspects. First, our \textbf{VQT} only uses the additional tokens to generate queries, not keys and values, for the MSA module. Thus, it does not change the intermediate features of a Transformer at all. In contrast, the additional tokens in VPT generate queries, keys, and values, and thus can be queried by other tokens and change their intermediate features. Second, and more importantly, while our \textbf{VQT} leverages the corresponding outputs of the additional tokens as summarized intermediate features, VPT in its Deep version disregards such output features entirely. \emph{In other words, these two methods take fundamentally different routes to approach transfer learning: \textbf{VQT} learns to leverage the existing intermediate features, while VPT aims to adapt the intermediate features.} 
As will be demonstrated in \autoref{exp}, these two routes have complementary strengths and can be compatible to further unleash the power of transfer learning.
It is worth noting that most of the recent methods towards parameter-efficient transfer learning (PETL), such as Prefix Tuning~\cite{li2021prefix} and AdaptFormer~\cite{chen2022adaptformer}, all can be considered adapting the intermediate features~\cite{he2021towards}. Thus, the aforementioned complementary strengths still apply.

%It is worth noting that most of the recent methods towards parameter-efficient transfer learning (PETL), such as Adapter~\cite{houlsby2019parameter}, Prefix Tuning~\cite{li2021prefix}, AdaptFormer~\cite{chen2022adaptformer}, etc., all can be considered as adapting the intermediate features~\cite{he2021towards}. Thus, the aforementioned difference from VQT applies to them as well. As will be demonstrated in \autoref{experiment: compatible}, VQT can be easily combined with these approaches to further unleash the power of transfer learning.

\begin{figure}
\centering
  \begin{subfigure}{\linewidth}
    \includegraphics[width=\linewidth]{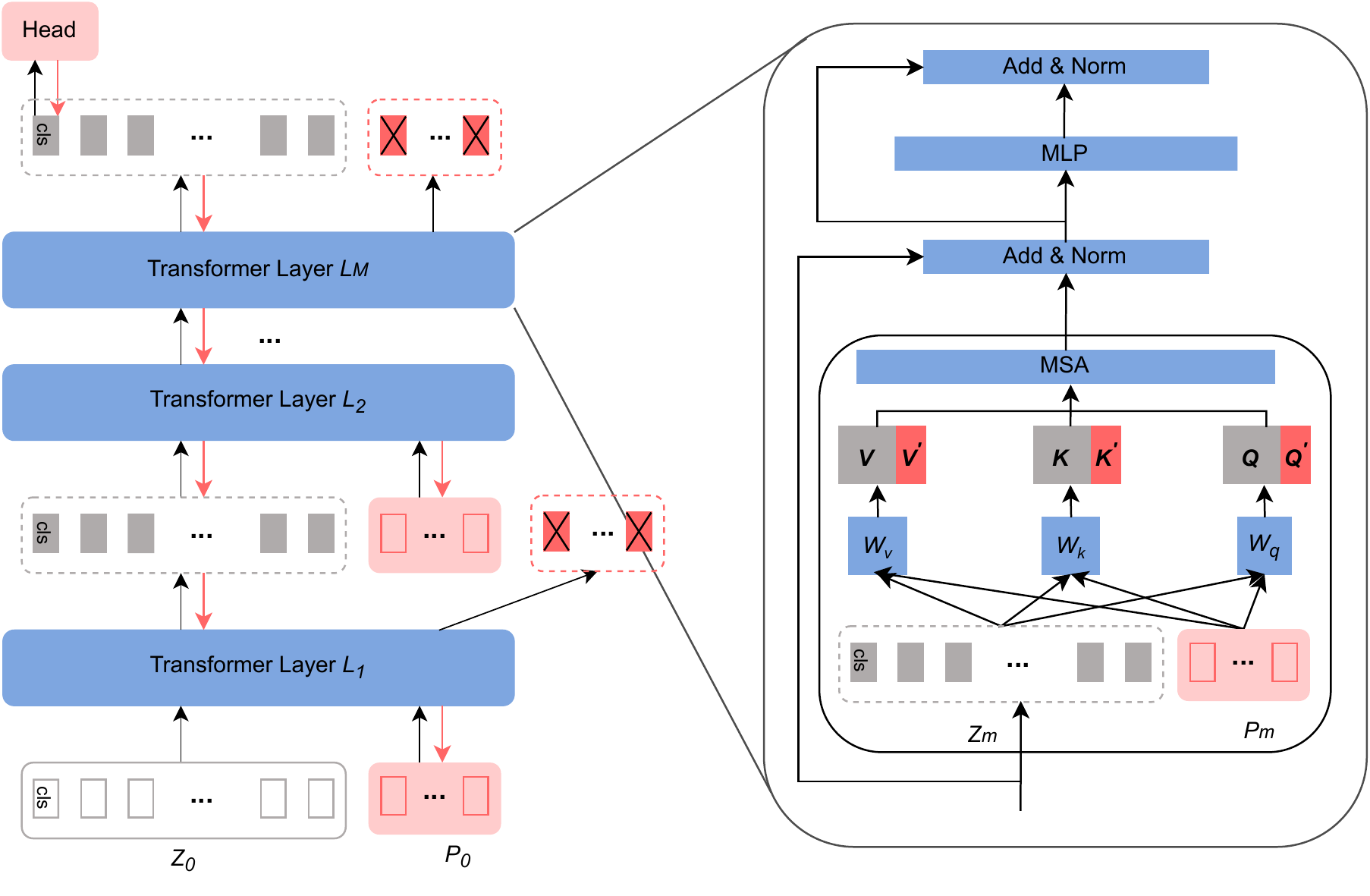}
    \vskip -5pt
    \caption{VPT: Visual Prompt Tuning (deep version)~\cite{jia2022vpt}}
    \vskip 5pt
    \label{fig:vpt}
  \end{subfigure}
  \begin{subfigure}{\linewidth}
    \includegraphics[width=\linewidth]{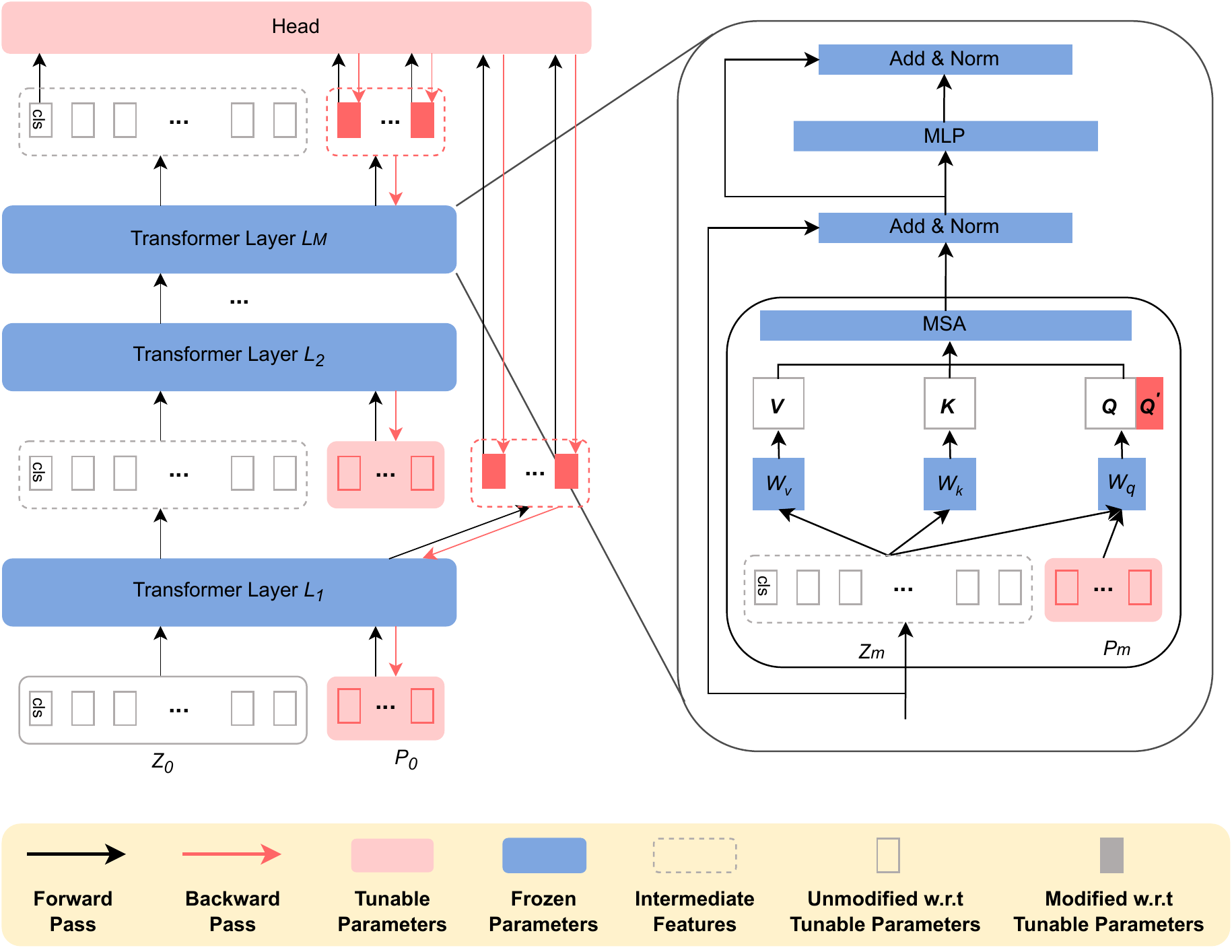}
    \caption{\textbf{Our VQT: Visual Query Tuning}}
    \label{fig:vqt}
  \end{subfigure}
\vskip-5pt
  \caption{\textbf{Our Visual Query Tuning (VQT) vs. Visual Prompt Tuning (VPT)~\cite{jia2022vpt}.} Our \textbf{VQT} allows \textit{linear probing} to directly access the intermediate features of a frozen Transformer model for parameter-efficient transfer learning. The newly introduced {\color{red}query tokens} in \textbf{VQT} (marked by the {\color{red}red} empty boxes in the {\color{red}red} shaded areas) only append additional columns (\ie, {\color{red}$\mQ'$}) to the Query features $\mQ$, not to 
  the Value features $\mV$ and the Key features $\mK$.  Thus, VQT keeps the {\color{Gray}intermediate features} intact ({\color{Gray} gray} empty boxes), enabling it to bypass expensive back-propagation steps in training (hence memory efficient). In contrast, VPT modifies the {\color{Gray}intermediate features} ({\color{Gray} gray} solid boxes) and needs more memory to learn its prompts. Please see~\autoref{approach} for details.}
  %\caption{The newly added query tokens from VQT do not impact the Value features $\mV$ and the Key features $\mK$, and only append additional columns {\color{blue}$\mQ'$} to $\mQ$. Thus VQT keeps all the output features intact while VPT modifies the output features by the inserted tokens. VPT requires storing intermediate back-propagation results to update the added tokens, while VQT bypasses expensive back-propagation steps since intermediate features are not modified}
 \label{fig: vqt_vpt}
 \vskip-10pt
\end{figure}

Besides the difference in how to approach transfer learning, another difference between \textbf{VQT} and many other PETL methods, including VPT, is memory usage in training. While many of them freeze (most of) the backbone model and only learn to adjust or add some parameters, the fact that the intermediate features are updated implies the need of a full back-propagation throughout the backbone, which is memory-heavy. In contrast, \textbf{VQT} keeps all the intermediate features intact and only learns to combine them. Learning the query tokens thus bypasses many paths in the standard back-propagation, reducing the memory footprint by 76\% compared to VPT.

We validate \textbf{VQT} on various downstream visual recognition tasks, using a pre-trained ViT~\cite{dosovitskiy2021an} as the backbone. \textbf{VQT} surpasses the SOTA method that utilizes intermediate features~\cite{evci2022head2toe} and \emph{full fine-tuning} in most tasks. We further demonstrate the robust and mutually beneficial compatibility between \textbf{VQT} and existing PETL approaches using different pre-trained backbones, including self-supervised and image-language pre-training. Finally, \textbf{VQT} achieves much higher accuracy than other PETL methods in a low-memory regime, suggesting that it is a more memory-efficient method.  
%The performance gain obtained by combing \textbf{VQT} and PETL methods cannot be achieved by simply increasing the complexity of PETL methods. 

% Moreover, VQT is robust to different pre-training setups. On self-supervised pre-trained backbones, VQT notably outperforms other baselines.
% We further demonstrate the compatibility of VQT with existing PEFL approaches, leading to a performance gain that cannot be achieved simply by increasing the parameters of these approaches.

%\zheda{which one should we use}
%Finally, \textbf{VQT} is able to reduce more than 70\% of the training memory while still keep a competitive performance.
%\zheda{OR}\textbf{VQT} achieves much higher accuracy than other PETL methods in a low-memory regime, suggesting that \textbf{VQT} is a more memory-efficient method. 

To sum up, our key contributions are 
\begin{enumerate} [noitemsep,topsep=0pt,parsep=0pt,partopsep=0pt,leftmargin=*]
    \item We propose \textbf{VQT} to aggregate intermediate features of Transformers for effective linear probing, featuring parameter and memory efficient transfer learning. %This allows us significantly to reduce the training memory usage and provide additional information to train a new prediction head for the downstream head. 
    %\item \textbf{VQT} is complementary to most of the recently proposed PETL methods that update the pre-trained backbone to generate new feature embeddings.
    \item \textbf{VQT} is compatible with other PETL methods that adapt intermediate features, further boosting the performance.
    \item \textbf{VQT} is robust to different pre-training setups, including self-supervised and image-language pre-training. 
\end{enumerate}

\section{Related Work}

% \WLC{I took a quick pass, looks nice. I will skip it for now and move to the approach part.}

% \zheda{i will add citations later once you finish editing}
%\subsection{Transformer}

\mypara{Transformer.} The splendent success of Transformer models~\cite{vaswani2017attention} in natural language processing (NLP)~\cite{wolf2020transformers} has sparked a growing interest in adopting these models in vision and multi-modal domains~\cite{khan2022transformers}. Since the proposal of the Vision Transformer (ViT)~\cite{dosovitskiy2021an}, Transformer-based methods have demonstrated impressive advances in various vision tasks, including image classification~\cite{liu2021swin, yu2022metaformer, touvron2021training}, image segmentation~\cite{xie2021segformer, strudel2021segmenter}, object detection~\cite{zhu2020deformable, carion2020end}, video understanding~\cite{arnab2021vivit, liu2022video}, point cloud processing~\cite{guo2021pct, zhao2021point}, and several other use cases~\cite{ chen2021pre, yang2020learning}. As Transformer models assume minimal prior knowledge about the structure of the problem, they are often pre-trained on large-scale datasets~\cite{chen2021empirical, caron2021emerging, radford2021learning}. Given that the Transformer models are notably larger than their convolutional neural network counterparts, e.g., ViT-G (1843M parameters)~\cite{zhai2022scaling} vs. ResNet-152 (58M parameters)~\cite{he2016deep}, how to adapt the pre-trained Transformers to downstream tasks in a parameter and memory efficient way remains a crucial open problem. 

%To this end, we propose a simple yet effective approach, Visual Query Tuning (VQT), to leverage frozen intermediate features of Transformer models for parameter and memory efficient transfer learning.

\mypara{PETL.}
The past few years have witnessed the huge success of parameter-efficient transfer learning (PETL) in NLP, aiming to adapt large pretrained language models (PLMs) to downstream tasks~\cite{kenton2019bert, brown2020language}. Typically, PETL methods insert small learnable modules into PLMs and fine-tune these modules with downstream tasks while freezing the pretrained weights of PLMs~\cite{he2021towards,lester2021power,he2022hyperprompt,mao2022unipelt,sung2021training,zaken2022bitfit,asai2022attentional,vu2022spot,liu2022ptuning,su2022transferability,zhong2022panda}. The current dominance of Transformer models in the vision field has urged the development of PETL methods in ViT~\cite{jia2022vpt,chen2022adaptformer,jie2022convolutional,zhang2022neural,liu2022polyhistor,lian2022scaling}. Recently, Visual Prompt Tuning (VPT)~\cite{jia2022vpt} was proposed to prepend learnable prompts to the input embeddings of each Transformer layer. 
AdaptFormer~\cite{chen2022adaptformer} inserts a bottleneck-structured fully connected layers parallel to the MLP block in a Transformer layer. 
Convpass~\cite{jie2022convolutional} inserts a convolutional bottleneck module while NOAH~\cite{zhang2022neural} performs a neural architecture search on existing PETL methods. 
Unlike all the aforementioned methods that update the output features of each Transformer layer, our VQT focuses on leveraging the frozen intermediate features. Thus, VQT is compatible with most existing PETL methods and enjoys memory efficiency.

%This module consists of two fully connected layers and a nonlinear activation function. 
%These prompts interact with and update the patch embeddings with self-attention during the fine-tuning process. 

\mypara{Transfer learning with intermediate features.}
Intermediate features of a pre-trained model contain rich and valuable information, which can be leveraged in various tasks such as object detection~\cite{lin2017feature, bell2016inside, hariharan2015hypercolumns} and OOD detection~\cite{lee2018simple}, etc. Recently, multiple works~\cite{dalvi2020analyzing, evci2022head2toe, eom2022layover, sung2022lst} have demonstrated the effectiveness of these features on transfer learning. On the NLP side, LST~\cite{sung2022lst} trains a lightweight Transformer network that takes intermediate features as input and generates output features for predictions. On the CV side, Evci~\etal~\cite{evci2022head2toe} attribute the success of fine-tuning to the ability to leverage intermediate features and proposed Head2Toe to select features from all layers for efficient transfer learning. Eom~\etal~\cite{eom2022layover} proposed utilizing intermediate features to facilitate transfer learning for multi-label classification. However, due to the massive number of intermediate features, most methods rely on the pooling operation to reduce the dimensionality, which may distort or eliminate useful information. This observation motivates us to introduce VQT, which learns to summarize intermediate features according to the downstream task.

%as the pooling operation is not learnable to choose relevant features according to the downstream task. 
%through the self-attention mechanism.
% , GAN training~\cite{sauer2021projected}, 

\section{Approach}
\label{approach}

We propose \textbf{Visual Query Tuning (VQT)} to adapt pre-trained Transformers to downstream tasks while keeping the backbone frozen. VQT keeps all the intermediate features intact and only learns to ``summarize'' them for \emph{linear probing} by introducing learnable ``query'' tokens to each layer. 
% We first provide the preliminaries and notations in \autoref{approach: preliminary}, followed by the formal description of VQT in \autoref{approach: vqt}. %Finally in \autoref{approach: compare}, we discuss the connection of VQT to related methods.
% % \autoref{sec: mem}
% We summarize this section by discussing the connection of VQT to related methods in \autoref{approach: compare}. 

%We propose visual query tuning (VQT) to adapt pre-trained Transformer models to downstream tasks while keeping the backbone frozen without any back-propagation through it. VQT keeps all the intermediate features intact and introduces learnable ``query'' tokens in each layer to summarize rich intermediate features for prediction head training. We first provide the preliminaries and notations in Sec.~\ref{approach: preliminary} followed by the formal description of VQT in Sec.~\ref{approach: vqt}. We will summarize this section by discussing VQT's connections with other related methods in Sec.~\ref{approach: compare}. 

%%---------------------------------------------------------------------------------------------------------------------------------------------------------------

\subsection{Preliminaries}
\label{approach: preliminary}
\subsubsection{Vision Transformer}

Vision Transformers (ViT)~\cite{dosovitskiy2021an} adapt the Transformer-based models~\cite{vaswani2017attention} from NLP into visual tasks, by dividing an image $\mI$ into a sequence of $N$ fixed-sized patches $\{\mI^{(n)}\}_{n=1}^N$ and treating them as NLP tokens. Each patch $\mI^{(n)}$ is first embedded into a $D$-dimensional vector $\vx^{(n)}_0$ with positional encoding. The sequence of vectors is then prepended with a ``CLS'' vector $\vx_0^{(\text{Class})}$ to generate the input $\mZ_0 = [\vx_0^{(\text{Class})}, \vx^{(1)}_0, \cdots, \vx^{(N)}_0]\in\R^{D\times(1+N)}$ to the ViT. We use superscript/subscript to index token/layer.

Normally, a ViT has $M$ layers, denoted by $\{L_m\}_{m=1}^M$. Given the input $\mZ_0$, the first layer $L_1$ generates the output $\mZ_1 = L_1(\mZ_0)=[\vx_1^{(\text{Class})}, \vx^{(1)}_1, \cdots, \vx^{(N)}_1]\in\R^{D\times(N+1)}$, which is of the same size as $\mZ_0$. That is, $\mZ_1$ has $1+N$ feature tokens, and each corresponds to the same column in $\mZ_0$. 
Such layer-wise processing then continues to generate the output of the next layer, $\mZ_m = L_m(\mZ_{m-1})$ for $m = 2, \cdots, M$, taking the output of the previous layer as input. Finally, the ``CLS'' vector $\vx_M^{(\text{Class})}$ in $\mZ_M$ is used
as the feature for prediction. Taking classification as an example, the predicted label $\hat{y} = \cst{Head}(\vx_M^{(\text{Class})})$ is generated by a linear classifier (\ie, a fully-connected layer).

\mypara{Details of each Transformer layer.} Our approach takes advantage of the inner workings of Transformer layers. In the following, we provide a concise background.

Each Transformer layer consists of a Multi-head Self-Attention (MSA) block, a Multi-Layer Perceptron (MLP) block, and several other operations including layer normalization and residual connections. Without loss of generality, let us consider a single-head self-attention block and disregard those additional operations.

Given the input $\mZ_{m-1}$ to $L_{m}$, the self-attention block first projects it into three matrices, namely Query $\mQ_m$, Key $\mK_m$, and Value $\mV_m$,
\begin{align}
\mQ_m = \mW_q \mZ_{m-1}, \quad \mK_m = \mW_k \mZ_{m-1}, \quad \mV_m = \mW_v \mZ_{m-1}. \label{eq_project}
\end{align}
Each of them has $1+N$ columns\footnote{For brevity, we ignore the layer index $m$ for the projection matrices $\mW_q, \mW_k, \mW_v$, but each layer has its own projection matrices.}, corresponding to each column (\ie, token) in $\mZ_{m-1}$. Then, the output of $L_{m}$, \ie, $\mZ_{m}$, can be calculated by:
\begin{align}
    & \mZ_{m} = \cst{MLP}_{m}\circ\cst{MSA}_{m}(\mZ_{m-1}), \\
    & \text{where} \hspace{5pt}
    \cst{MSA}_{m}(\mZ_{m-1}) = \mV_m\times\cst{Softmax}(\frac{\mK_m^\top\mQ_m}{\sqrt{D}}). \label{eq_MSA}
\end{align}
The $\cst{Softmax}$ is taken over elements of each column; the $\cst{MLP}_{m}$ is applied to each column of $\cst{MSA}_{m}(\mZ_{m-1})$ independently.

\subsubsection{Transfer Learning: Linear Probing, Fine-tuning, and Intermediate Feature Utilization}
To adapt a pre-trained ViT to downstream tasks, \emph{linear probing} freezes the whole backbone model but the prediction head: it disregards the original $\cst{Head}$ and learns a new one. \emph{Fine-tuning}, on top of \emph{linear probing}, allows the backbone model to be updated as well.

%\subsubsection{Intermediate Feature Utilization}
Several recent works have demonstrated the effectiveness of utilizing intermediate features in transfer learning, by allowing \emph{linear probing} to directly access them~\cite{evci2022head2toe, eom2022layover}. The seminal work \HT~\cite{evci2022head2toe} takes intermediate features from $\mZ_0$ and four distinct steps in each Transformer layer: features after the layer normalization, after the MSA block, and inside and after the MLP block. Since each of them has $1+N$ tokens, \HT groups tokens by their indices and performs average pooling to reduce the dimensionality. The resulting features --- over each group, step, and layer --- are then concatenated together for \emph{linear probing}.
To further reduce dimensionality, \HT 
employs group lasso~\cite{yuan2006model, argyriou2006multi} for feature selection.

We note that while the second dimensionality reduction is driven by downstream tasks, the first (\ie, pooling) is not, which may inadvertently eliminate useful information. This shortcoming motivates us to develop Visual Query Tuning (VQT) for the effective usage of intermediate features.

%Without changing anything parts in the backbone, the seminar work Head2Toe selects features $\mathbf{h}_l$ at the $l$ layer and processes the features with a fixed function $a_l(\cdot)$. The concatenation of the processed features is used for the new prediction head training. To determine the relevant features for the new task, Head2Toe employs group lasso for feature selection. However, Head2Toe relies on average pooling in $a_l(\cdot)$ to reduce the dimensionality of plenteous intermediate features, which may eliminate useful information because the pooling operation is static and not learnable to choose relevant features for downstream tasks. This observation motivates us to introduce Visual Query Tuning, which leverages the intrinsic mechanism of Transformer models to summarize rich intermediate features in each layer. 

% The feature process function $a_l(\cdot)$ in Head2Toe performs average pooling to reduce dimensionality and L2 normalization to preserve the relative magnitude of features within a layer. However, 

\iffalse
\begin{align}
 \mathbf{h}=\left[a_1\left(\mathbf{h}_1\right), \ldots, a_L\left(\mathbf{h}_L\right)\right]
\end{align}
\fi

%%---------------------------------------------------------------------------------------------------------------------------------------------------------------

\subsection{Visual Query Tuning (VQT)}
\label{approach: vqt}

We propose to replace the average pooling operation in \HT with the intrinsic ``summarizing'' mechanism in Transformers. We note that the MSA block introduced in~\autoref{eq_MSA} essentially performs weighted averages of the Value features $\mV$ over tokens, in which the weights are determined by the columns of $\mK^\top\mQ$. {That is, if we can append additional ``columns'' to $\mK^\top\mQ$, the MSA block will output additional weighted combinations of $\mV$.} In the special case that the appended vector to $\mK^\top\mQ$ 
has identical entries (\eg, an all-zero vector), the weighted average reduces to a simple average. In other words, average pooling can be thought of as a special output of the MSA layer.

Taking this insight into account, we propose to learn and append additional columns {\color{blue}$\mQ'$} to $\mQ$. We realize this idea by introducing a handful of $T$ learnable ``query'' tokens $\mP_{m-1} = [\vp_{m-1}^{(1)}, \cdots, \vp_{m-1}^{(T)}]$ to the input of each Transformer layer $L_{m}$. See~\autoref{fig:vqt} for an illustration. Different from the original input $\mZ_{m-1}$ that undergoes the three projections introduced in~\autoref{eq_project}, $\mP_{m-1}$ only undergoes the projection by $\mW_q$,
\begin{align}
    \mQ'_m = \mW_q\mP_{m-1}.
\end{align}
By appending {\color{blue}$\mQ'_m$} to $\mQ_m$ column-wise, we modify the computation of the original MSA block in \autoref{eq_MSA} by
\begin{align}
& \mV_m\times\cst{Softmax}(\frac{\mK_m^\top[\mQ_m, {\color{blue}\mQ'_m}]}{\sqrt{D}}) = \label{eq_MSA_new}\\
& [\mV_m\times\cst{Softmax}(\frac{\mK_m^\top\mQ_m}{\sqrt{D}}), {\color{blue}\mV_m\times\cst{Softmax}(\frac{\mK_m^\top\mQ'_m}{\sqrt{D}})}].\nonumber 
\end{align}
The second half ({\color{blue}blue} color) corresponds to the newly summarized MSA features by the learnable query tokens $\mP_{m-1}$. 
Then after the MLP block $\cst{MLP}_{m}$, these features lead to the newly summarized features $\mZ'_m\in\R^{D\times T}$ from layer $L_m$. 
We can then concatenate these newly summarized features over layers, $\mZ'_m\in\R^{D\times T}$ for $m = 1, \cdots, M$, together with the final ``CLS'' vector $\vx_M^{(\text{Class})}$, for \emph{linear probing}. We name our approach \textbf{Visual Query Tuning (VQT)}, reflecting the fact that the newly added tokens $\mP_{m}$ for $m = 0, \cdots, M-1$ only serve for the additional columns in Query matrices.

\mypara{Properties of VQT.} As indicated in~\autoref{eq_MSA_new}, the newly introduced query tokens do not change the MSA features the pre-trained ViT obtains (\ie, the first half). This implies that VQT keeps all the original intermediate features (\eg, $\mZ_m$) intact but only learns to recombine them.

\mypara{Training of VQT.} Given the training data of the downstream task, the query tokens $\{\mP_m\}_{m=0}^{M-1}$ are learned end-to-end with the new prediction head, which directly accesses the outputs $\{\mZ'_{m+1}\}_{m=0}^{M-1}$ of these query tokens.

To further reduce the dimensionality of $\{\mZ'_{m+1}\}_{m=0}^{M-1}$, we optionally employ group lasso, following \HT \cite{evci2022head2toe}. In detail, we first learn the query tokens without group lasso. We then freeze them and apply group lasso to select useful features from $\{\mZ'_{m+1}\}_{m=0}^{M-1}$. We also explored various ways for dimension reduction in Appendix~\ref{suppl-sec:variants}.

%%%%%%%%%%%%%%%%%%%%%%%%%%%%%%%%%%%%%%%%%%%%%%%%%%%

%Unlike most existing transfer learning methods that focus on adapting the model to produce new features, we propose Visual Query Tuning (VQT) that keeps all the features intact and summarizes the intermediate features for predictions. Concretely, we introduce a set of learnable embeddings \textit{(prompts)} to the input of every Transformer layer when we generate the Query vector $\mathbf{Q}$ and therefore, we call these prompts \textit{Query Prompts}. For the $l+1$ layer, the input is $\mZ_l = \left[\mathbf{x}_l, \mathbf{E}_l\right] \in \mathbb{R}^{(N+1)\times d}$ and by inserting the query prompts $\mathbf{P}_l=\left\{\mathbf{p}_l^j \in \mathbb{R}^d \mid  1\leq j \leq p\right\}$, we can generate a new input $\tilde{\mZ_l} = \left[\mathbf{P}_l, \mathbf{x}_l, \mathbf{E}_l\right] \in \mathbb{R}^{(p+N+1)\times d}$ where $p$ is number of query prompts. The MSA block of VQT, $ {MSA}^{VQT}$ is formulated as:

\subsection{Comparison to Related Works}
\label{approach: compare}

\mypara{Comparison to \HT~\cite{evci2022head2toe}.} We list several key differences between \HT and our VQT. First, compared to \HT, which takes intermediate features from multiple steps in a Transformer layer, VQT only takes the newly summarized intermediate features after each layer. Second, and more importantly, VQT employs a different way to combine intermediate features \emph{across tokens}. Generally speaking, there are two ways to combine a set of feature vectors $\{\vx^{(n)}\in\R^D\}_{n=1}^N$: concatenation and average pooling. The former assumes that different vectors have different meanings even at the same dimension, which is suitable for features across layers. The latter assumes that the same dimension means similarly to different vectors so they can be compared and averaged, which is suitable for features across tokens.
One particular drawback of the former is the dimensionality (\ie, inefficiency). For the latter, it is the potential loss of useful information since it combines features blindly to the downstream tasks (\ie, ineffectiveness). 
\HT takes a mix of these two ways to combine features over tokens, and likely suffers one (or both) drawbacks. In contrast, VQT 
leverages the intrinsic mechanism of self-attention to aggregate features adaptively, conditioned on the features and the learnable query tokens, making it a more efficient and effective way to tackle the numerous intermediate features within each layer. 
 
\mypara{Comparison to Visual Prompt Tuning (VPT).} At first glance, VQT may be reminiscent of VPT~\cite{jia2022vpt}, but they are fundamentally different as highlighted in~\autoref{sec:intro} and \autoref{fig: vqt_vpt}. Here, we provide some more details and illustrations. 

VPT in its deep version (VPT-Deep) introduces learnable tokens $\mP_{m-1} = [\vp_{m-1}^{(1)}, \cdots, \vp_{m-1}^{(T)}]$ to the input of each Transformer layer $L_{m}$, similarly to VQT. However, unlike VQT which uses $\mP_{m-1}$ only for querying, VPT-Deep treats $\mP_{m-1}$ the same as other input tokens $\mZ_{m-1}$ and generates the corresponding Query, Key, and Value matrices,
\begin{align}
\mQ'_m = \mW_q \mP_{m-1}, \quad \mK'_m = \mW_k \mP_{m-1}, \quad \mV'_m = \mW_v \mP_{m-1}. \nonumber  %\label{eq_project_2}
\end{align}
These matrices are then appended to the original ones from $\mZ_{m-1}$ (cf.~\autoref{eq_project}) before self attention, 
\begin{align}
\tilde{\mQ}_m = [\mQ_m, \mQ'_m], \hspace{4pt} \tilde{\mK}_m = [\mK_m, \mK'_m], \hspace{4pt} \tilde{\mV}_m = [\mV_m, \mV'_m], \nonumber %\label{eq_project_2}
\end{align}
making the output of the MSA block as
\begin{align}
& \tilde{\mV}_m \times\cst{Softmax}(\frac{\tilde{\mK}_m^\top\tilde{\mQ}_m}{\sqrt{D}}) =  \label{eq_MSA_2} \\
& [\tilde{\mV}_m \times\cst{Softmax}(\frac{\tilde{\mK}_m^\top\mQ_m}{\sqrt{D}}), \tilde{\mV}_m \times\cst{Softmax}(\frac{\tilde{\mK}_m^\top\mQ'_m}{\sqrt{D}})]. \nonumber
\end{align}
Compared to \autoref{eq_MSA} and \autoref{eq_MSA_new}, the first half of the matrix in \autoref{eq_MSA_2} changes, implying that all the intermediate features as well as the final ``CLS'' vector $\vx_M^{(\text{Class})}$ are updated according to the learnable tokens $\mP_{m-1}$. In contrast, VQT keeps these (intermediate) features intact.
% corresponding to the original tokens $\mZ_{m-1}$
 
Perhaps more subtly but importantly, VPT-Deep ends up dropping the second half of the matrix in \autoref{eq_MSA_2}. In other words, VPT-Deep does not exploit the newly summarized features by $\mQ'_m$ at all, making it conceptually similar to Prefix Tuning~\cite{li2021prefix}.
Please see~\autoref{fig: vqt_vpt} for a side-by-side comparison between VQT and VPT-Deep.

The aforementioned differences suggest an interesting distinction between VQT and VPT: \emph{VQT learns to leverage the existing intermediate features, while VPT learns to adapt the intermediate features.} In~\autoref{experiment: compatible}, we demonstrate one particular strength of VQT, which is to transfer self-supervised pre-trained models.

\mypara{Comparison and Compatibility with PETL methods.} In fact, most of the existing PETL approaches that adjust or add a small set of parameters to the backbone model update the intermediate features~\cite{he2021towards}. Thus, our VQT is likely complementary to them and can be used to boost their performance. In~\autoref{experiment: compatible}, we explore this idea by introducing learnable query tokens to these methods. 

\mypara{Memory efficiency in training.}
As pointed out in~\cite{sung2022lst}, when learning the newly added parameters, most PETL methods require storing intermediate back-propagation results, which is memory-inefficient for large Transformer-based models. For VQT, since it keeps all the intermediate features intact and only learns to (i) tune the query tokens (ii) and \emph{linearly probe} the corresponding outputs of them, the training bypasses many expensive back-propagation paths, significantly reducing the memory footprint. See~\autoref{experiment: memory} for details.

%Moreover, as pointed out in~\cite{sung2022lst}, although most existing PETL methods, including VPT, freeze the backbone model, updating the added parameters requires storing intermediate backpropagation results, which is memory-inefficient for large Transformer-based models. Contrastingly, since VQT keeps all the intermediate features intact and only utilizes the outputs of the query prompts for prediction, updating query prompts does not require expensive backpropagation of the large backbone Transformer, which significantly cuts down the memory footprint. As shown in Sec.~\ref{experiment: memory}, VQT reduces the memory consumption by XXX\% compared to VPT-Deep. 

% \begin{figure}[tb]
%     \centering
%     \includegraphics[width=0.6\linewidth]{CVPR 23/figures/petl.png}
%     \caption{bla}
%     \label{fig:petl}
% \end{figure}

\section{Experiments}
\label{exp}
\begin{table*}[t!]
\normalsize
\resizebox{\textwidth}{!}{
\begin{tabular}{l|ccccccc:c|cccc:c|cccccccc:c|c}
\toprule
& \multicolumn{8}{c|}{Natural}&\multicolumn{5}{c|}{Specialized}&\multicolumn{9}{c}{Structured}&\\
Method& {\rotatebox[origin=l]{90}{CIFAR-100}}&{\rotatebox[origin=l]{90}{Caltech101}}&{\rotatebox[origin=l]{90}{DTD}}&{\rotatebox[origin=l]{90}{Flowers102}}&{\rotatebox[origin=l]{90}{Pets}}&{\rotatebox[origin=l]{90}{SVHN}}&{\rotatebox[origin=l]{90}{Sun397}}&{\rotatebox[origin=l]{90}{Mean}}&{\rotatebox[origin=l]{90}{Camelyon}}&{\rotatebox[origin=l]{90}{EuroSAT}}&{\rotatebox[origin=l]{90}{Resisc45}}&{\rotatebox[origin=l]{90}{Retinopathy}}&{\rotatebox[origin=l]{90}{Mean}}&{\rotatebox[origin=l]{90}{Clevr-Count}}&{\rotatebox[origin=l]{90}{Clevr-Dist}}&{\rotatebox[origin=l]{90}{DMLab}}&{\rotatebox[origin=l]{90}{KITTI-Dist}}&{\rotatebox[origin=l]{90}{dSpr-Loc}}&{\rotatebox[origin=l]{90}{dSpr-Ori}}&{\rotatebox[origin=l]{90}{sNORB-Azim}}&{\rotatebox[origin=l]{90}{sNORB-Elev}}&{\rotatebox[origin=l]{90}{Mean}}&{\rotatebox[origin=l]{90}{Overall Mean}}\\
\midrule
Scratch&7.6&19.1&13.1&29.6&6..7&19.4&2.3&14.0&71.0&71.0&29.3&72.0&60.8&31.6&52.5&27.2&39.1&66.1&29.7&11.7&24.1&35.3&32.8\\
Linear-probing&50.6&85.6&61.4&79.5&86.5&40.8&38.0&63.2&79.7&91.5&71.7&65.5&77.1&41.4&34.4&34.1&55.4&18.1&26.4&16.5&24.8&31.4&52.7 \\
Fine-tuning&44.3&84.5&54.1&84.7&74.7&\textbf{87.2}&26.9&65.2&\textbf{85.3}&95.0&76.0&70.4&81.7&\textbf{71.5}&\textbf{60.5}&\textbf{46.9}&72.9&\textbf{74.5}&38.7&28.5&23.8&\textbf{52.2}&63.2 \\
\midrule

\HT&54..4&86.8&64.1&83.4&82.6&78.9&32.1&68.9&81.3&95.4&81.2&73.7&82.9&49.0&57.7&41.5&64.4&52.3&32.8&\textbf{32.7}&\textbf{39.7}&46.3&62.3\\

\textbf{VQT (Ours)} &\textbf{58.4}&\textbf{89.4}&\textbf{66.7}&\textbf{90.4}&\textbf{89.1}&81.1&\textbf{33.7}&\textbf{72.7}&82.2&\textbf{96.2}&\textbf{84.7}&\textbf{74.9}&\textbf{84.5}&50.8&57.6&43.5&\textbf{77.2}&65.9&\textbf{43.1}&24.8&31.6&49.3&\textbf{65.3}\\

\bottomrule
\end{tabular}}
\vskip-10pt
\caption{Test accuracy on the VTAB-1k benchmark with ViT-B/16 pre-trained on ImageNet-1K. "Mean" denotes the average accuracy for each category and "Overall Mean" shows the average accuracy over 19 tasks.}
\label{table:h2t}
\vskip-5pt
\end{table*}

\subsection{Experiment Setup} \label{experiment: exp_setup}
%\zheda{we can shorten this apart if we dont have enough space}

\mypara{Dataset.} 
We evaluate the transfer learning performance on the \textbf{VTAB-1k}~\cite{zhai2019large}, which consists of 19 image classification tasks categorized into three groups: Natural, Specialized, and Structured. The Natural group comprises natural images captured with standard cameras. The Specialized group contains images captured by specialist equipment for remote sensing and medical purpose. The Structured group evaluates the scene structure comprehension, such as object counting and 3D depth estimation.
Following~\cite{zhai2019large}, we perform an 80/20 split on the \textbf{1000} training images in each task for hyperparameter searching. The reported result (top-1 classification accuracy) is obtained by training on the 1000 training images and evaluating on the original test set. 
%More details are included in Appendix.  

% The classes may represent generic, fine-grained, or abstract objects. 

\mypara{Pre-training setup.} We use ViT-B/16~\cite{dosovitskiy2021an} as the backbone. The pre-training setup follows the corresponding compared baselines. When comparing with Head2Toe, we use ImageNet-1K supervised pre-trained backbone. When investigating the compatibility with other PETL methods, ImageNet-21K supervised pre-trained backbone is used. To demonstrate the robustness of VQT to different pre-training setups, we also evaluate VQT on self-supervised (MAE)~\cite{he2022masked} and image-language (CLIP) pre-trained \cite{radford2021learning} backbones. Please see Appendix~\ref{suppl-sec:exp_details} for more details.

%We include more details about the pre-training methods and backbones in Appendix.

\begin{table}[t]
    \footnotesize
    \centering
    \begin{tabular}{llll}
        \toprule
        Methods & Natural & Specialized & Structured \\
        \midrule
        \multicolumn{4}{c}{CLIP backbone}\\
        \midrule
AdaptFormer     & 82.6  &85.1 &60.9  \\
AdaptFormer+VQT & 82.1 $\textcolor{green}{0.5} \; \textcolor{green}\downarrow$&85.8 $\textcolor{red}{0.7} \; \textcolor{red}\uparrow$&62.6 $\textcolor{red}{1.7} \; \textcolor{red}\uparrow$\\
VPT             & 80.4 &84.9 &50.9  \\
VPT+VQT         & 81.5 $\textcolor{red}{1.1} \; \textcolor{red}\uparrow$&86.3 $\textcolor{red}{1.4} \; \textcolor{red}\uparrow$&57.2 $\textcolor{red}{6.3} \; \textcolor{red}\uparrow$\\
\midrule
\multicolumn{4}{c}{MAE backbone}\\
\midrule
AdaptFormer     & 68.7 &81.3 &58.3 \\
AdaptFormer+VQT & 71.1 $\textcolor{red}{2.4} \; \textcolor{red}\uparrow$&83.3 $\textcolor{red}{2.0} \; \textcolor{red}\uparrow$&59.2 $\textcolor{red}{0.9} \; \textcolor{red}\uparrow$\\
VPT             & 63.5 &79.1 &48.6 \\
VPT+VQT         &67.9 $\textcolor{red}{4.4} \; \textcolor{red}\uparrow$&82.7 $\textcolor{red}{3.6} \; \textcolor{red}\uparrow$&49.7 $\textcolor{red}{1.1} \; \textcolor{red}\uparrow$\\
\midrule
\multicolumn{4}{c}{Supervised ImageNet-21K backbone}\\
\midrule
AdaptFormer     & 80.1 &82.3 &50.3   \\
AdaptFormer+VQT & 79.6 $\textcolor{green}{0.5} \; \textcolor{green}\downarrow$&84.3 $\textcolor{red}{2.0} \; \textcolor{red}\uparrow$&53.0   $\textcolor{red}{2.7} \; \textcolor{red}\uparrow$\\
VPT             & 79.1 &84.6 &54.4   \\
VPT+VQT         & 78.9 $\textcolor{green}{0.2} \; \textcolor{green}\downarrow$&83.7 $\textcolor{green}{0.9} \; \textcolor{green}\downarrow$&54.6 $\textcolor{red}{0.2} \; \textcolor{red}\uparrow$  \\

% \midrule
%         \multicolumn{4}{c}{Supervised ImageNet-1K backbone}\\
%     \midrule
%     AdaptFormer     & 73.4&80.1&47.3\\
%     AdaptFormer+VQT & 75.4 $\textcolor{red}{2.0} \; \textcolor{red}\uparrow$&84.5 $\textcolor{red}{4.4} \; \textcolor{red}\uparrow$&53.1 $\textcolor{red}{5.8} \; \textcolor{red}\uparrow$\\
%     VPT& 75.5&81.2&45.2\\
%     VPT+VQT&75.4 $\textcolor{green}{0.1} \; \textcolor{green}\downarrow$&84.9 $\textcolor{red}{3.7} \; \textcolor{red}\uparrow$&50.5 $\textcolor{red}{5.3} \; \textcolor{red}\uparrow$\\
% \bottomrule
        \bottomrule
    \end{tabular}
    \vskip-10pt
    \caption{\textbf{Compatibility of VQT} with AdaptFormer and VPT on MAE, CLIP, and supervised pre-trained backbones.}
    \vskip-10pt
    \label{tab:petl}
\end{table}

% \begin{figure}[tb]
%     \centering
%     \includegraphics[width=\linewidth]{CVPR 23/figures/compatibility_old.png}
%     \caption{\textbf{Compatibility of VQT} with AdaptFormer and VPT on MAE, CLIP and supervised pre-trained backbones.}
%     \label{fig:petl}
% \end{figure}

% \begin{figure}[tb]
%     \centering
%     \includegraphics[width=\linewidth]{CVPR 23/figures/compatibility_new.pdf}
%     \caption{directly using prompts in vpt. VQT's superiority on MAE and CLIP diminishes, vpt+vqt is worse on Natural CLIP, Specialized and Structured Supervised
% }
%     \label{fig:petl}
% \end{figure}

\subsection{Effectiveness of VQT}
\label{experiment: effectiveness}
To evaluate the transfer learning performance of VQT, we compare VQT with methods that fix the whole backbone (\emph{linear-probing} and \HT) and full \emph{fine-tuning}, which updates all network parameters end to end. For a fair comparison, \textbf{we match the number of tunable parameters in VQT with that in \HT} (details are included in Appendix~\ref{suppl-sec:main_compare_h2t}). In general, VQT improves over \emph{linear-probing} by 12.6\% and outperforms \HT and full \emph{fine-tuning} by 3\% and 2.1\% respectively, on average performance over 19 tasks, which demonstrates \textbf{the strength of using intermediate features and the effectiveness of VQT in summarizing them}. In the {Natural} category, VQT surpasses \HT and \emph{fine-tuning} by 2.8\% and 7.5\%, respectively, and outperforms them in the {Specialized} category by 1.6\% and 2.8\%, respectively. As shown in~\cite{evci2022head2toe, zhai2019large}, the Natural and Specialized categories have stronger domain affinities with the source domain (ImageNet) since they are all real images captured by cameras. Thus, the pre-trained backbone can generate more relevant intermediate features for similar domains. The only exception is the {Structured} category consisting of rendered artificial images from simulated environments, which differs significantly from ImageNet. Although VQT continues to improve \HT, \emph{fine-tuning} shows 2.9\% enhancement over VQT, suggesting that if we need to adapt to a more different targeted domain, we may consider tuning a small part of the backbone to produce updated features for new data before applying our VQT techniques. Appendix~\ref{suppl-sec:more_comp_to_h2t} contains more comparisons between \HT and VQT.

% To evaluate if VQT can summarize intermediate features better than average pooling for downstream tasks, we compare VQT with \HT on the VTAB-1k benchmark using ViT-B/16 pretrained on ImageNet-2012, following the setup in \HT. As shown in~\autoref{table:h2t}, VQT improves over \emph{linear-probing} by 12.6\% and outperforms \HT and full \emph{fine-tuning} by around 2\% on average performance over 19 tasks, which demonstrate the effectiveness of VPT in summarizing intermediate features. In the Natural category, VQT surpasses \HT and \emph{fine-tuning} by 1.4\% and 7.5\%, respectively, and outperforms them in the Specialized category by 3.1\% and 1.9\%, respectively. However, in the Structured category, although VQT continues to improve \HT, \emph{fine-tuning} shows 2.9\% enhancement over VQT. This phenomenon could be explained by the domain affinities between the source and target domains. As shown in~\cite{evci2022head2toe, zhai2019large}, the Natural and Specialized categories have stronger domain affinities with the source domain (ImageNet) since they are all real images captured by cameras. On the contrary, the Structured category consists of images generated from simulated environments, which differs significantly from ImageNet. Therefore, it suggests that if we need to adapt to a more different targeted domain, we may consider tuning a small part of the backbone to produce updated features for new data before applying our VQT techniques. 

\subsection{Compatibility with PETL Methods in Different Pre-training Methods}
\label{experiment: compatible}
As mentioned in \autoref{approach: compare}, most existing PETL methods and VQT take fundamentally different routes to approach transfer learning: PETL methods focus on adapting the model to generate updated features, while VQT aims to better leverage features. Building upon this conceptual complementariness, we investigate if they can be combined to unleash the power of transfer learning. Moreover, in order to demonstrate the robustness of the compatibility, we evaluate performance on three different pre-trained backbones: self-supervised pre-trained (MAE with ImageNet-1K)~\cite{he2022masked}, image-language pre-trained (CLIP)~\cite{radford2021learning} and supervised pre-trained (ImageNet-21K).

Specifically, we focus on two recently proposed methods:  AdaptFormer~\cite{chen2022adaptformer} and VPT~\cite{jia2022vpt}. AdaptFormer inserts fully connected layers in a bottleneck structure parallel to the MLP block in each Transformer layer~\cite{chen2022adaptformer}; VPT
adds learnable tokens to the input of every Transformer layer. 

To equip AdaptFormer~\cite{chen2022adaptformer} and VPT~\cite{jia2022vpt} with our VQT, firstly, we update the pre-trained model with AdaptFormer or VPT so that the model can generate relevant intermediate features for the downstream task. Then we add \textbf{$T=1$} query token to the input of every layer to summarize the updated intermediate features. For AdaptFormer, we use the default bottleneck dimension 64; for VPT, we use the best number of added tokens for each task reported in their paper. 

% To equip AdaptFormer with our VQT, firstly, we update the pretrained model with AdaptFormer so that model can generate relevant intermediate features for the downstream task. Then we add one query token to the input of every layer to summarize the updated intermediate features. For VPT, we directly leverage the output of their added tokens, which otherwise will be discarded, as the summary of the intermediate features in each layer.  For AdaptFormer, we use the default bottleneck dimension 64 and we use the best number of added tokens for VPT reported in their paper. 

\begin{figure}[tb]
    \centering
    \includegraphics[width=0.6\linewidth]{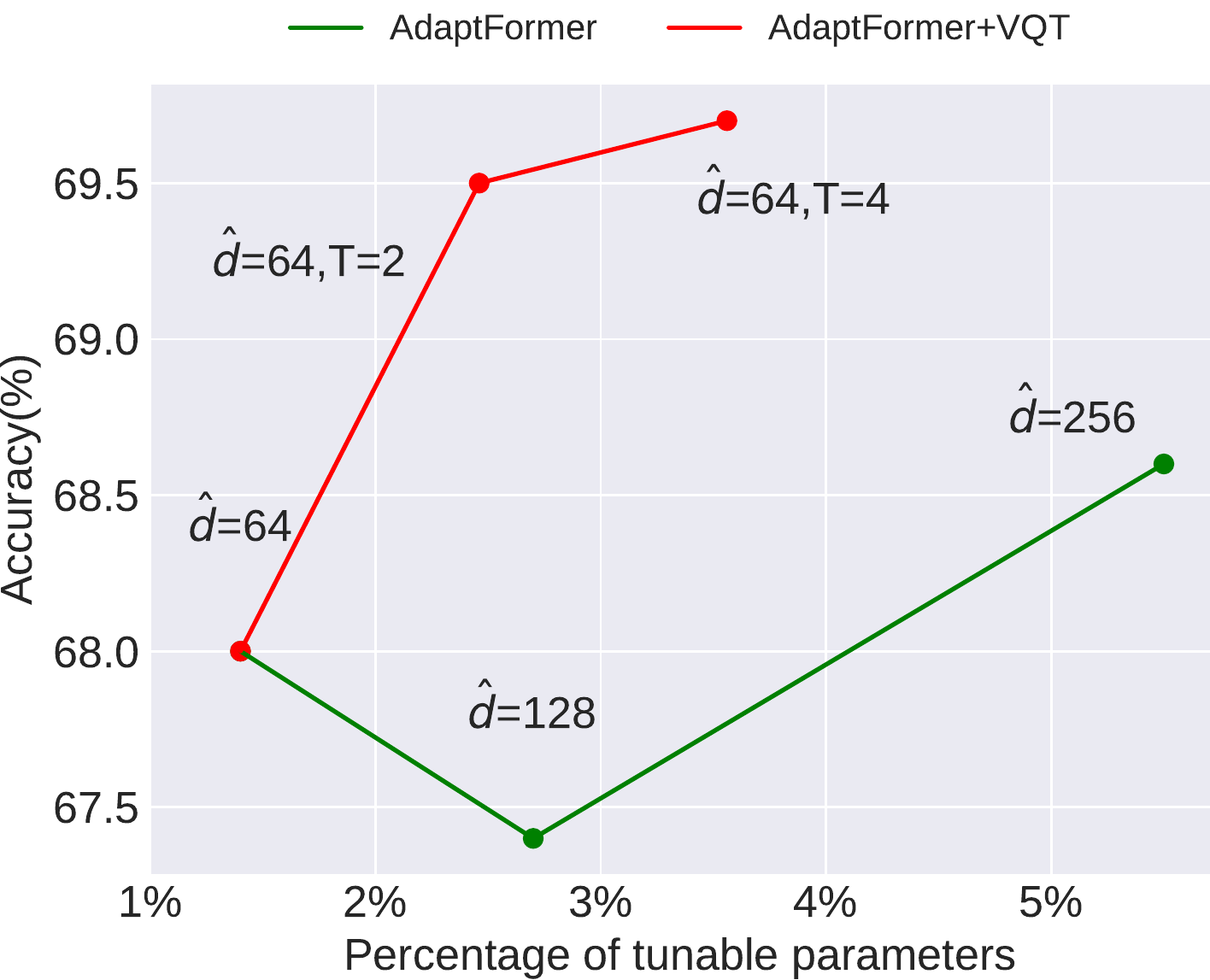}
        \vskip-5pt
    \caption{The power of leveraging intermediate features provided by VQT allows AdaptFormer to incorporate additional information from the updated model ({\color{red}red curve}), which would not be possible by simply increasing the complexity of the inserted modules ({\color{green}green curve}). $\hat{d}$ denotes the bottleneck dimension of AdaptFormer and T represents the number of VQT's query tokens.}
    \label{fig:adapt_vqt}
\end{figure}

We summarize the results in \autoref{tab:petl}, where each row shows the results for one pre-trained backbone, and each column shows the results for one data category. Generally speaking, AdaptFormer and VPT benefit from VQT in most of the scenarios across different data categories and pre-trained backbones. \textbf{The improvement is more salient in the MAE backbone.} Since the MAE pre-training uses the reconstruction objective instead of the classification or contrastive one, we hypothesize that some useful intermediate features for classification may not be propagated to the final layer\footnote{\autoref{tab:mae} shows the transfer learning results by each method alone, using the MAE backbone. Our VQT notably outperforms other methods.}. With the help of VQT, AdaptFormer and VPT can leverage intermediate features in a more concise and effective way. Additionally, \textbf{VQT also benefits from AdaptFormer and VPT.} In \autoref{experiment: effectiveness}, we found that directly applying VQT to the pre-trained backbone may not be effective for the Structured category due to the low domain affinity. With the intermediate features updated by AdaptFormer and VPT, VQT can summarize these more relevant features to improve the results for the Structured group. To sum up, the experiment results illustrate that \textbf{VQT and PETL methods are complementary and mutually beneficial}, with the potential to further unleash the power of transfer. We provide detailed results of various pre-trained backbones in Appendix~\ref{suppl-sec:results_backbones} and the compatibility comparison between \HT and VQT in Appendix~\ref{suppl-sec:comptibility}.

\begin{table}[t]
    \footnotesize
    \centering
    \begin{tabular}{c|ccc}
        \toprule
        Methods & Natural & Specialized & Structured \\
        \midrule
        Linear-probing & 18.87 & 53.72 & 23.70 \\
        Fine-tuning & 59.29 & 79.68 & 53.82 \\
        VPT         & 63.50 & 79.15 & 48.58 \\
        \textbf{VQT (Our)}         & 66.00 & 82.87 & 52.64 \\
        \bottomrule
    \end{tabular}
    \vskip-5pt
    \caption{Average accuracy on VTAB-1k using the MAE backbone.}
    \vskip-10pt
    \label{tab:mae}
\end{table}

% \WLC{In the experiments, when you show that ours is better, adding some hypothesis as follows. We hypothesize that self-supervised or webly-supervised pre-trained models may bury the most relevant features to supervised (downstream) tasks in the intermediate layers, as the models are not learned to prioritize those features to the final layer.}

To confirm that the improvement mentioned above does not simply come from the increase of tunable parameters, we enlarge AdaptFormer's added modules by increasing the bottleneck dimension $\hat{d}$ from 64 to 128 and 256 to match the tunable parameter number of AdaptFormer when it is equipped with VQT\footnote{For VPT, since we already use its best prompt sizes, adding more prompts to it will not improve its performance.}. As shown in \autoref{fig:adapt_vqt}, AdaptFormer with VQT significantly outperforms AdaptFormer with larger added modules when the numbers of tunable parameters are similar. 
This further demonstrates the complementary strength of VQT and AdaptFormer: the improvement by leveraging intermediate features summarized by VQT cannot be achieved by simply increasing the complexity of the inserted modules in AdaptFormer.

\begin{figure}
\centering
  \begin{subfigure}{0.49\linewidth}
    \includegraphics[width=\linewidth]{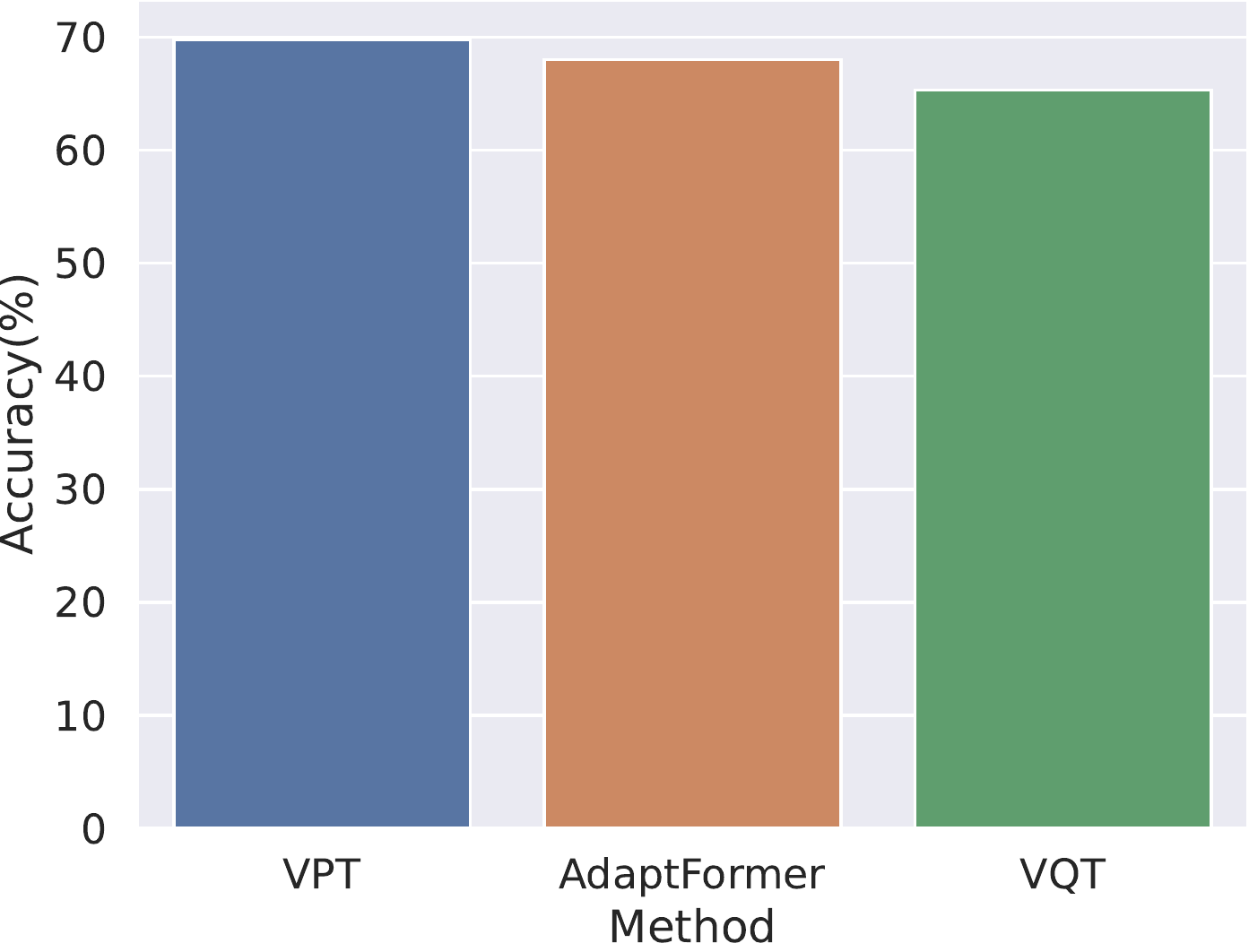}
    \caption{}
    \label{fig:best_perf}
  \end{subfigure}
  \begin{subfigure}{0.49\linewidth}
    \includegraphics[width=\linewidth]{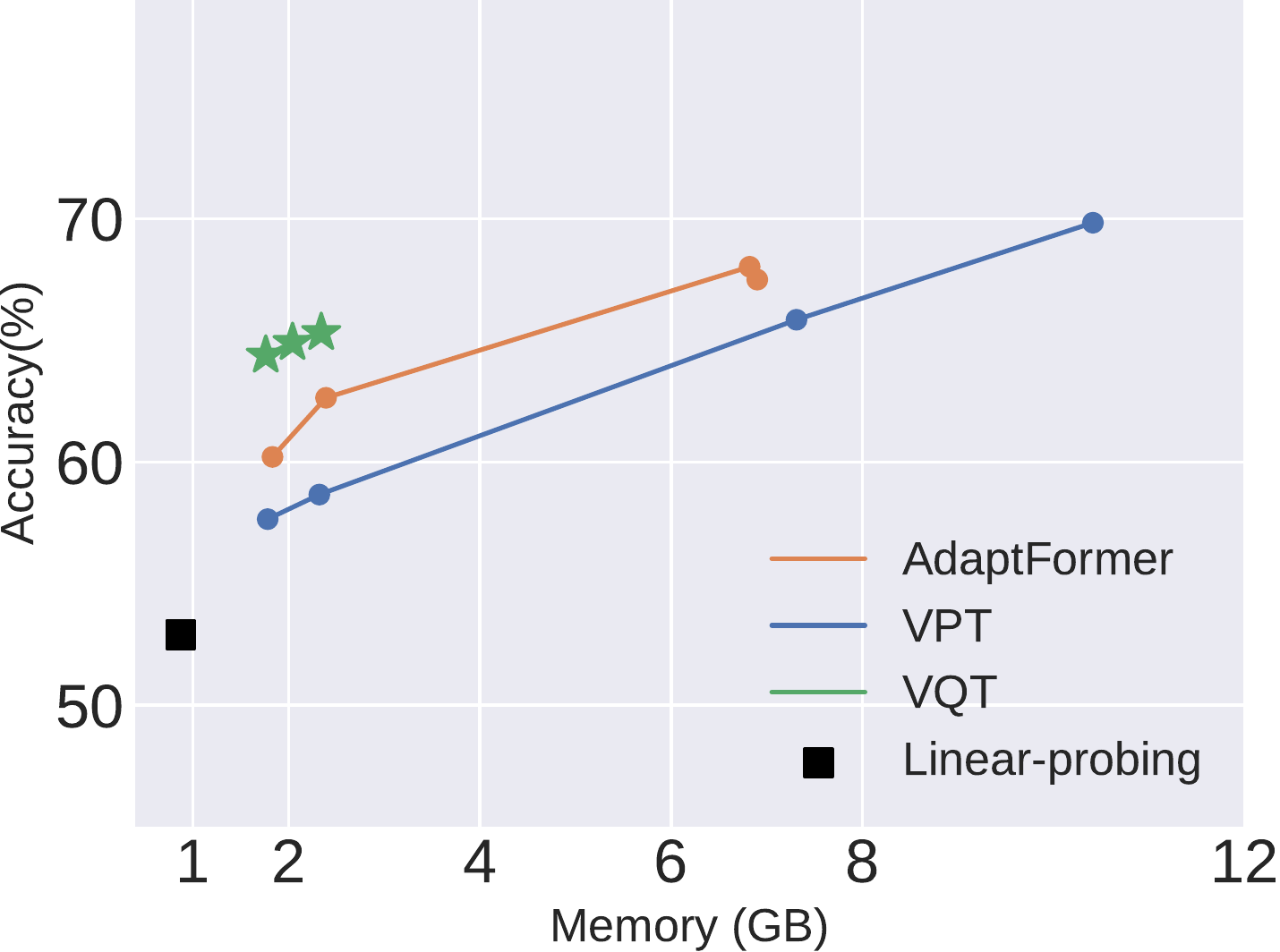}
    \caption{}
    \label{fig:mem}
  \end{subfigure}
  \vskip -10pt
    \caption{\textbf{Comparison under memory constraints.} (a) Without constraints, VPT and AdaptFormer slightly outperform VQT. (b) With constraints, VQT performs the best in low-memory regimes.}
    \vskip-10pt
  %\caption{(a) shows the performance comparison with ViT-B/16 pre-trained on ImageNet-21K. (b) shows the accuracy-memory trade-off for AdaptFormer, VPT and VQT.  VQT performs the best in low-memory regime. }
\end{figure}
%\mypara{Compatibility with other PETL methods} Most PETL methods focus on adapting the output features by inserting a few learnable modules into each Transformer layer. As shown in Figure.~\ref{fig:petl}, AdaptFormer inserts a bottleneck module parallel to the MLP block; ConvPass inserts a convolutional bottleneck module parallel to the MSA and the MLP block, and VPT prepends prompts to the patch embeddings. As our VQT addresses utilization of the intermediate features regardless of the features being updated or not, VQT is essentially compatible with existing PETL methods that update the output features. As demonstrated in Section.~\ref{experiment: compatible}, the power of leveraging intermediate features provided by VQT allows PETL methods to incorporate additional information from the updated model, which would not be possible by simply increasing the complexity of their inserted modules.  

% \subsection{VQT with Different Pretrained Setups}
% \begin{figure}[tb]
%     \centering
%     \includegraphics[width=1\linewidth]{CVPR 23/figures/mae_clip.pdf}
%     \caption{Adding VQP can help PETL methods more robust to pretrained setups, like SSL, CLIP}
%     \label{fig:mae_clip}
% \end{figure}

\subsection{Memory Efficient Training}
\label{experiment: memory}
While many PETL methods reduce the number of tunable parameters, they cannot cut down the memory footprint during training by much, and therefore, the evaluation of PETL methods often ignores 
%does not consider
memory consumption. In real-world scenarios, however, a model is often required to adapt to new data on edge devices for privacy concerns, necessitating the need for methods that can be trained with limited memory. This motivates us to further analyze the accuracy-memory trade-off for VPT, AdaptFormer, and VQT. 

As discussed in \autoref{approach: compare}, VPT and AdaptFormer require storing the intermediate back-propagation results to update their added parameters, while VQT bypasses the expensive back-propagation because it keeps all the intermediate features intact. To evaluate their performance in the low-memory regime, we only add their inserted parameters to the last few layers to match the memory usage. \autoref{fig:best_perf} shows the performance of VQT, VPT, and AdaptFormer under their \textbf{best hyperparameters without memory constraints}; \autoref{fig:mem} depicts the \textbf{accuracy-memory trade-off} for these methods. When memory is not a constraint, VPT and AdaptFormer slightly outperform VQT, but they consume 3.8x and 5.9x more memory (GB) than VQT, respectively, as we can see in \autoref{fig:mem}. 

When memory is a constraint (left side of \autoref{fig:mem}), we see drastic accuracy drops of AdaptFormer and VPT. Although they still surpass \emph{linear-probing}, VQT outperforms them significantly, suggesting that VQT is a more memory-efficient method thanks to its query-only mechanism.

% \begin{figure}[tb]
%     \centering
%     \includegraphics[width=0.75\linewidth]{CVPR 23/figures/mem.pdf}
%     \caption{Accuracy-memory trade off, if we want adaptformer and vpt to be more memory efficient, their performance drops significantly }
%     \label{fig:petl}
% \end{figure}

% \subsection{Visualization of Prompts}
% \begin{figure}[tb]
%     \centering
%     \includegraphics[width=1\linewidth]{CVPR 23/figures/t-sne.png}
%     \caption{VQT's t-sne should be more separable than linear, fin-tune and h2t}
%     \label{fig:petl}
% \end{figure}

\subsection{Discussion} \label{sec:discussion}
\begin{figure}[tb]
    \centering
    \includegraphics[width=\linewidth]{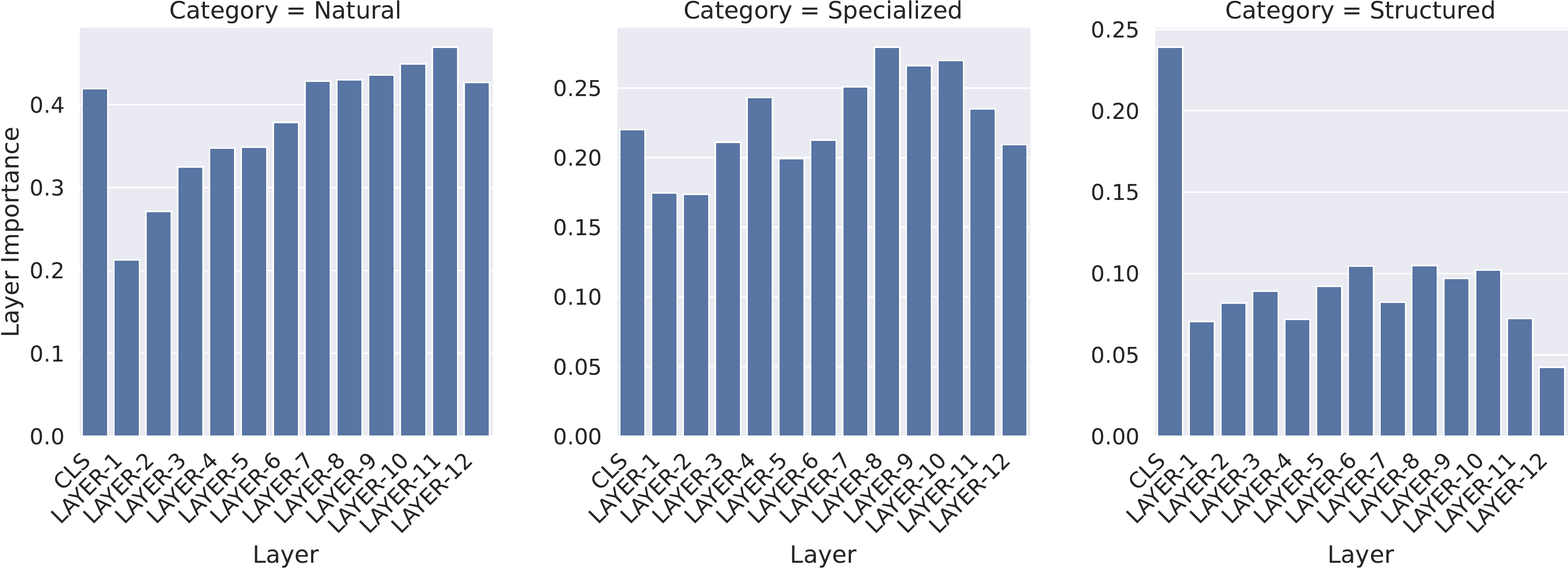}
        \vskip-10pt
    \caption{Layer importance for each category in VTAB-1k.}
    \label{fig:layer_impt}
    \vskip-5pt
\end{figure}

% \subsubsection{Variants of VQT}
% \begin{figure}[tb]
%     \centering
%     \includegraphics[width=1\linewidth]{CVPR 23/figures/variant.png}
%     \caption{5 variants: (1) sparsity, default (2) attention-across-layers (3) weighted-sum-across-layers (4) pooling-within-layer (5) weighted-sum-within-layer}
%     \label{fig:petl}
% \end{figure}
\mypara{Layer importance for each category.} 
As VQT leverages the summarized intermediate features for predictions, we investigate which layers produce more critical features for each category. 
In~\autoref{fig:layer_impt}, we show each layer's importance score computed by averaging the feature importance in the layer. 
%By averaging the feature importance in a layer, we show the layer importance in~\autoref{fig:layer_impt}. 
% and by aggregating the feature importance by layer, we show the layer importance in \autoref{fig:layer_impt}.
Features in deeper layers are more important for the Natural category, while features from all layers are almost equally important for the Specialized category. 
Contrastingly, VQT heavily relies on the CLS token for the Structured category. 
We hypothesize that the low domain affinity between ImageNet and the Structured category may cause the intermediate features to be less relevant, and the model needs to depend more on the CLS token. 

% As explained in \autoref{approach: vqt}, we apply group lasso for feature selection. 

%  because this category has lower domain affinity with the pretrain source domain; thus, the intermediate feature may not be relevant. 

\begin{figure}[tb]
    \centering
    \includegraphics[width=\linewidth]{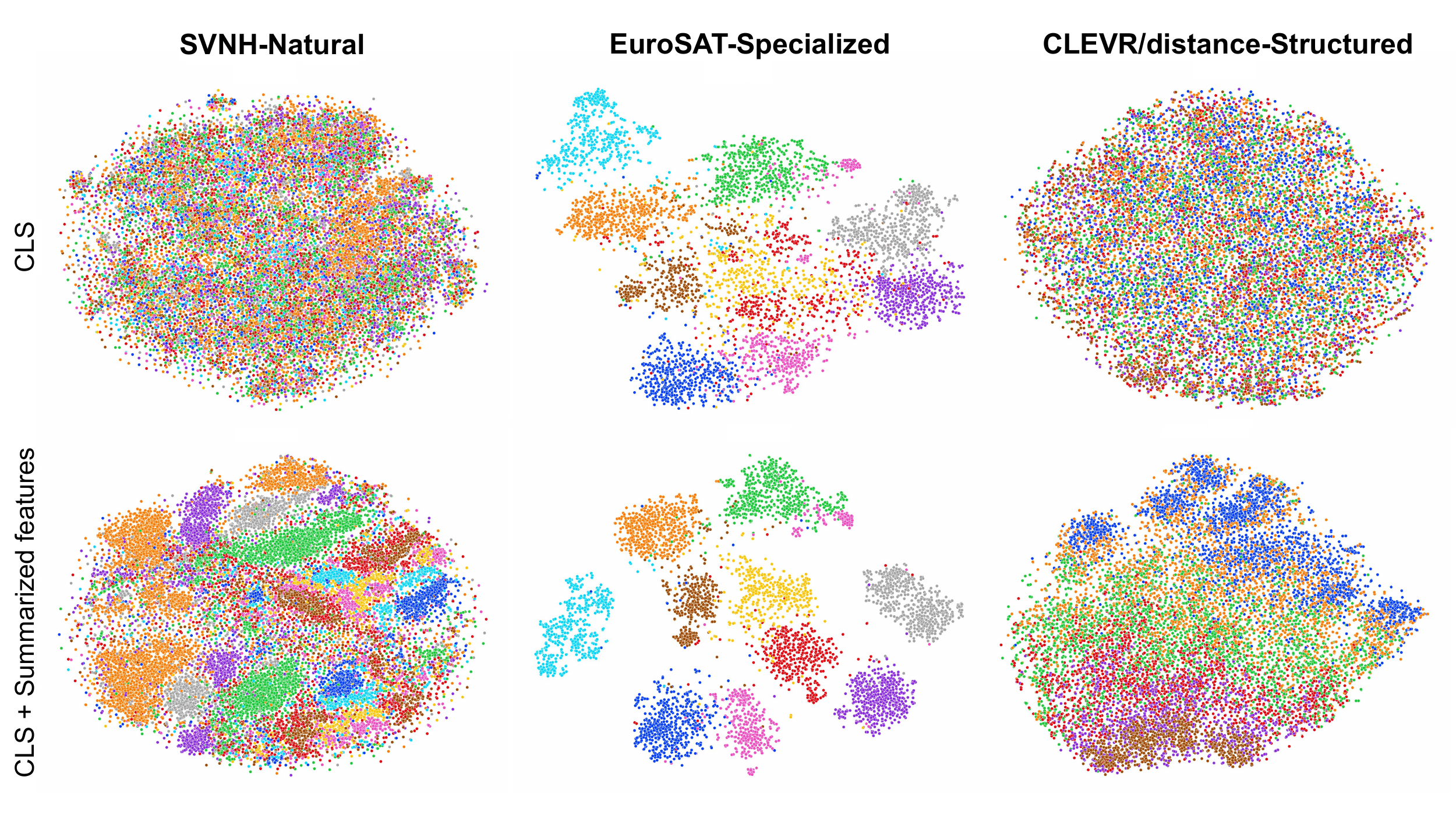}
    \vskip-10pt
    \caption{\textbf{t-SNE visualization of the CLS tokens alone (top) and CLS tokens plus our summarized features (bottom)} on 3 tasks from each VTAB's category. Adding the summarized intermediate features makes the whole features more separable. }
    \label{fig:tsne}
    \vskip-5pt
\end{figure}

\begin{figure}
\centering
  \begin{subfigure}{0.49\linewidth}
    \includegraphics[width=\linewidth]{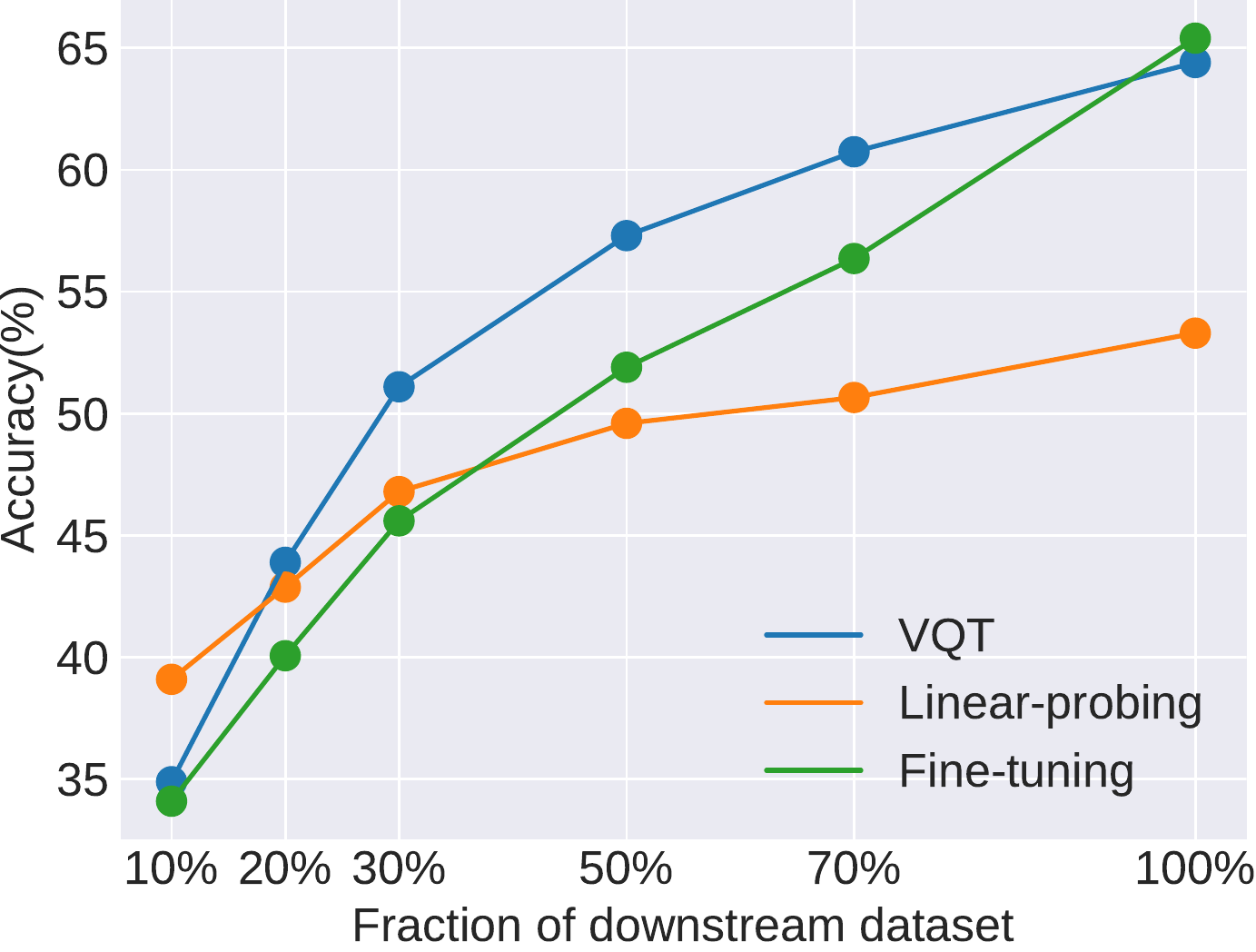}
    \caption{}
    \label{fig:frac_data}
  \end{subfigure}
  \begin{subfigure}{0.49\linewidth}
    \includegraphics[width=\linewidth]{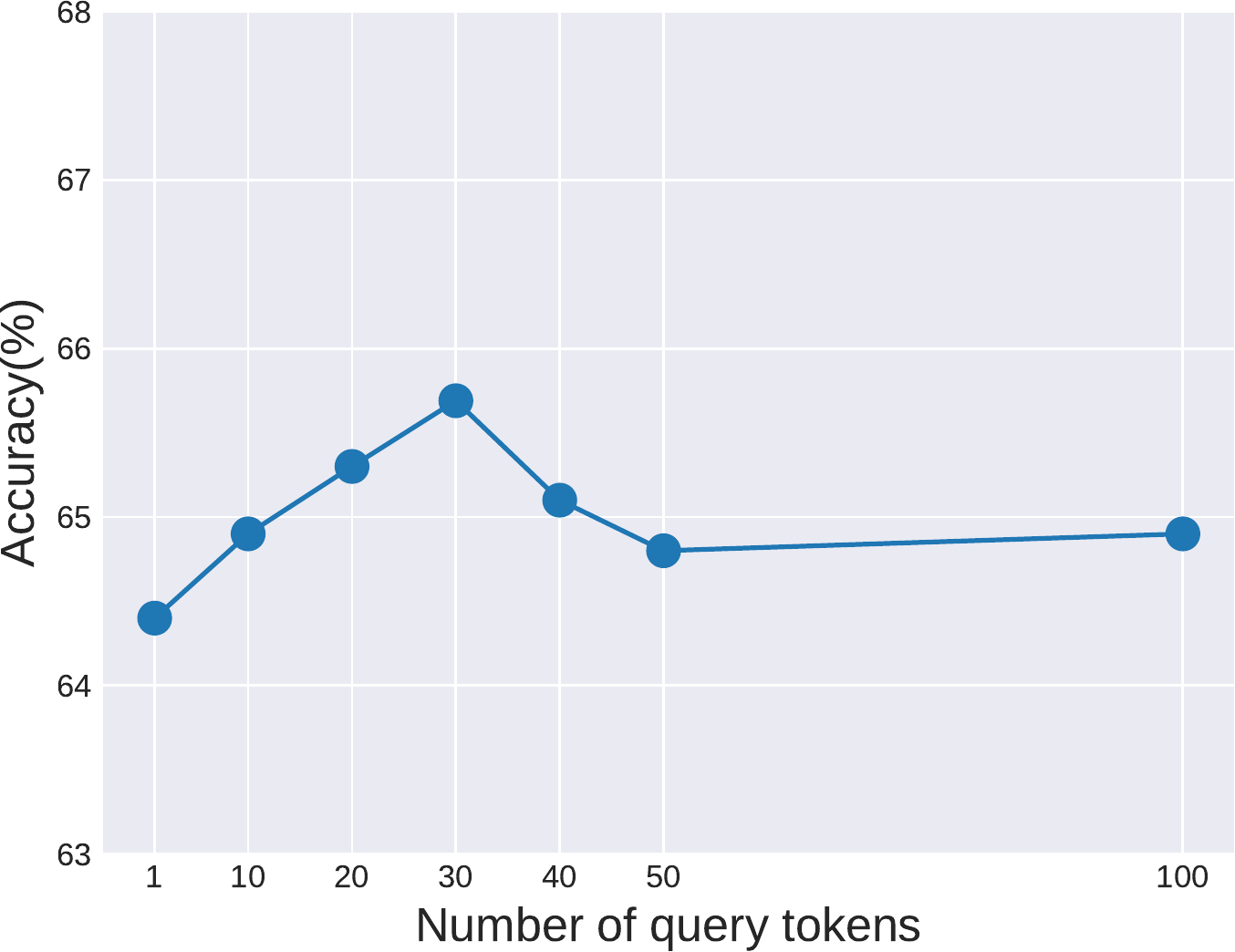}
    \caption{}
    \label{fig:q_num}
  \end{subfigure}
  \vskip-10pt
  \caption{(a) Average accuracy over the 19 tasks in VTAB-1k using different training data sizes. For each task, 100\% means that we use all the 1000 training images. In the 10\% data case, we averagely have only 2 images per class. (b) Average accuracy on VTAB-1k using different numbers of query tokens for VQT.}
  \vskip-10pt
\end{figure}

% \begin{figure}[tb]
%     \centering
%     \includegraphics[width=0.8\linewidth]{CVPR 23/figures/query_num.png}
%     \caption{Query number's impact on performance}
%     \label{fig:q_num}
% \end{figure}
\mypara{Different downstream data sizes.} 
We further study the effectiveness of VQT under various training data sizes. 
We reduce the VTAB's training sizes to \{10\%, 20\%, 30\%, 50\%, 70\%\} and compare VQT with Fine-tuning and Linear probing in~\autoref{fig:frac_data}.
Although fine-tuning slightly outperforms VQT on 100\% data, VQT consistently performs better as we keep reducing the training data. 
On the 10\% data case, where we only have 2 images per class on average, Linear probing obtains the best accuracy, but its improvement diminishes and performs much worse than VQT when more data become available. 
These results show that VQT is more favorable in a wide range of training data sizes. 

% We examine the impact of downstream data size on accuracy by reducing VTAB's training set size to \{10\%, 30\%, 50\%, 70\%\}. \emph{Linear-probing} outperforms \emph{fine-tuning} and VQT in extremely low-data regimes, 100 images/task ((10\%), but VQT consistently surpasses both of them once we have more than 200 images/task (20\%), illustrating that VQT is a data-efficient method. 

\mypara{Number of query tokens.}
%The number of query tokens is the only extra hyperparameter for VQT compared to fine-tuning. In all previous experiments, we use \emph{one} query token, and 
As we use only \emph{one} query token for VQT in previous experiments, we now study VQT's performance using more query tokens on VTAB-1k. 
% As the number of query tokens reflects VQT's model complexity, we study its impact on VTAB-1k. 
\autoref{fig:q_num} shows that more query tokens can improve VQT, but the accuracy drops when we add more than 40 tokens. 
We hypothesize that overly increasing the model complexity causes overfitting due to the limited data in VTAB-1k. 

\mypara{Visualization.}
\autoref{fig:tsne} shows t-SNE~\cite{van2008visualizing} visualization of the CLS token and our summarized features for three tasks (SVHN, EuroSAT, and Clevr-Dist), one from each category. Compared with the CLS token alone, adding summarized features makes the whole features more separable, showing the strength of using intermediate features and the effectiveness of our query tokens in summarizing them.  
We provide the visualization of other tasks in Appendix~\ref{suppl-sec:tsne}. 
% The visualization for other tasks are included in the Appendix.  

% \subsubsection{Prompt Depth}
% \begin{figure}[tb]
%     \centering
%     \includegraphics[width=0.8\linewidth]{CVPR 23/figures/prompt_depth.png}
%     \caption{Inserting prompts at different layers. Two x axes, bottom-up 1, 1-3, 1-6, 1-9, 1-12; top-down 1-12, 9-12, 6-12, 3-12. i-j means inserting prompts from i layer to j layer}
%     \label{fig:petl}
% \end{figure}

% \subsubsection{Different backbones}
% \begin{figure}[tb]
%     \centering
%     \includegraphics[width=0.8\linewidth]{CVPR 23/figures/backbone.png}
%     \caption{VQT's performance on different backbones, ViT B/L/H, x axis is the number of parameters}
%     \label{fig:petl}
% \end{figure}

\section{Conclusion}

We introduced Visual Query Tuning, a simple yet effective approach to aggregate intermediate features of Vision Transformers. By introducing a set of learnable ``query'' tokens to each layer, VQT leverages the intrinsic mechanism of Transformers to ``summarize'' rich intermediate features while keeping the intermediate features intact, which allows it to enjoy a memory-efficient training without back-propagation through the entire backbone. Empirically, VQT surpasses \HT, the SOTA method that utilizes intermediate features, and we demonstrate robust and mutually beneficial compatibility between VQT and other PETL methods. Furthermore, VQT is a more memory-efficient approach and achieves much higher performance in a low-memory regime. While VQT only focuses on summarizing features within each layer, we hope our work can pave the way for exploring more effective ways of using features across layers and leveraging intermediate features in transfer learning for other tasks, such as object detection, semantic segmentation and video classification.

{\small
\section*{Acknowledgments}
This research is supported in part by NSF (IIS-2107077, OAC-2118240, and OAC-2112606) and Cisco Research. We are thankful for the computational resources of the Ohio Supercomputer Center. We thank Yu Su (OSU) for the helpful discussion. We thank Menglin Jia and Luming Tang (Cornell) for code sharing.
}

%%%%%%%%% REFERENCES
{\small
\bibliographystyle{ieee_fullname}
\bibliography{egbib}
}

\clearpage
\appendix
\begin{figure*} [th!]
\centering
  \begin{subfigure}{0.15\textwidth}
    \includegraphics[width=\linewidth]{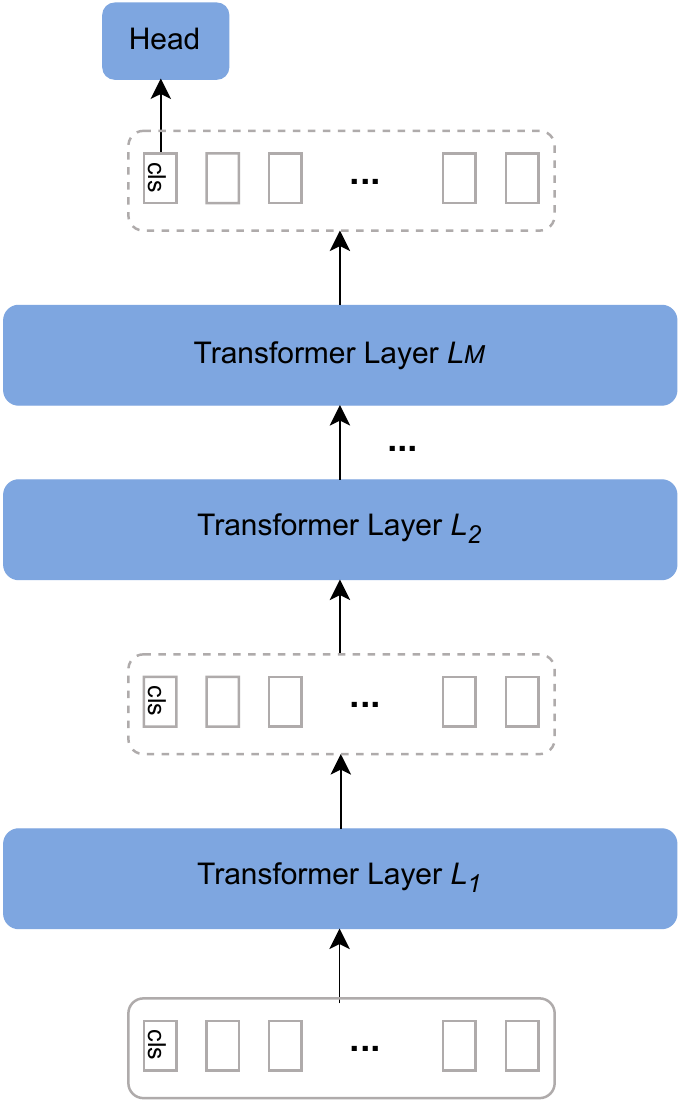}
    \caption{}
    \label{fig:vit}
  \end{subfigure}
  \hfill
  \begin{subfigure}{0.15\textwidth}
    \includegraphics[width=\linewidth]{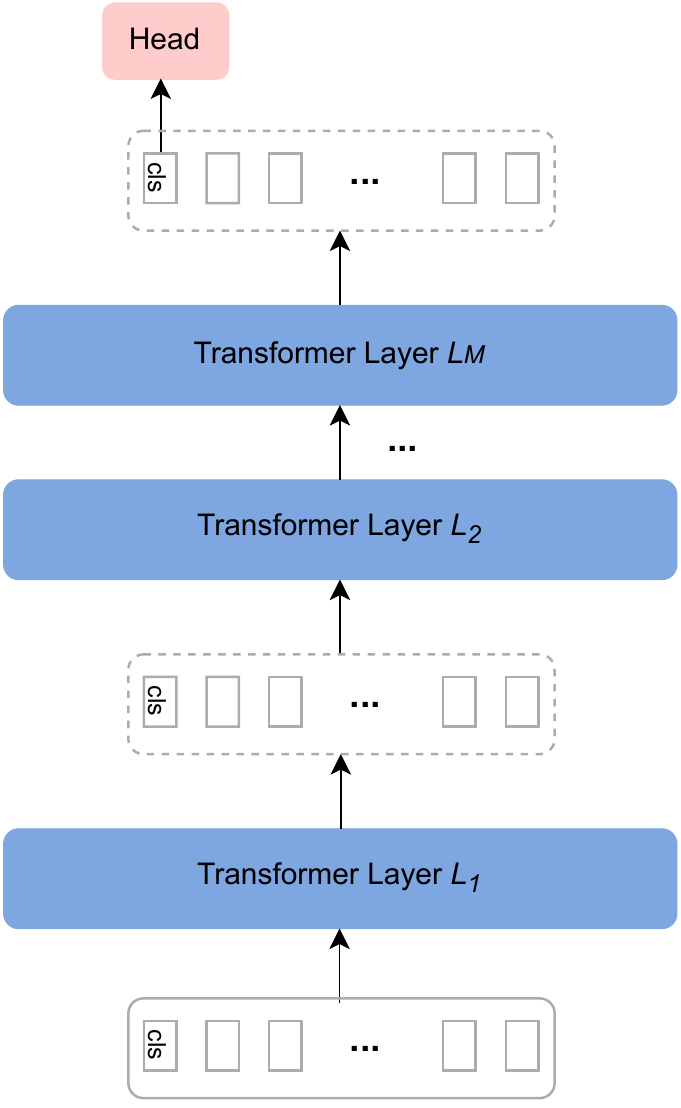}
    \caption{}
    \label{fig:linear}
  \end{subfigure}
  \hfill
    \begin{subfigure}{0.15\textwidth}
    \includegraphics[width=\linewidth]{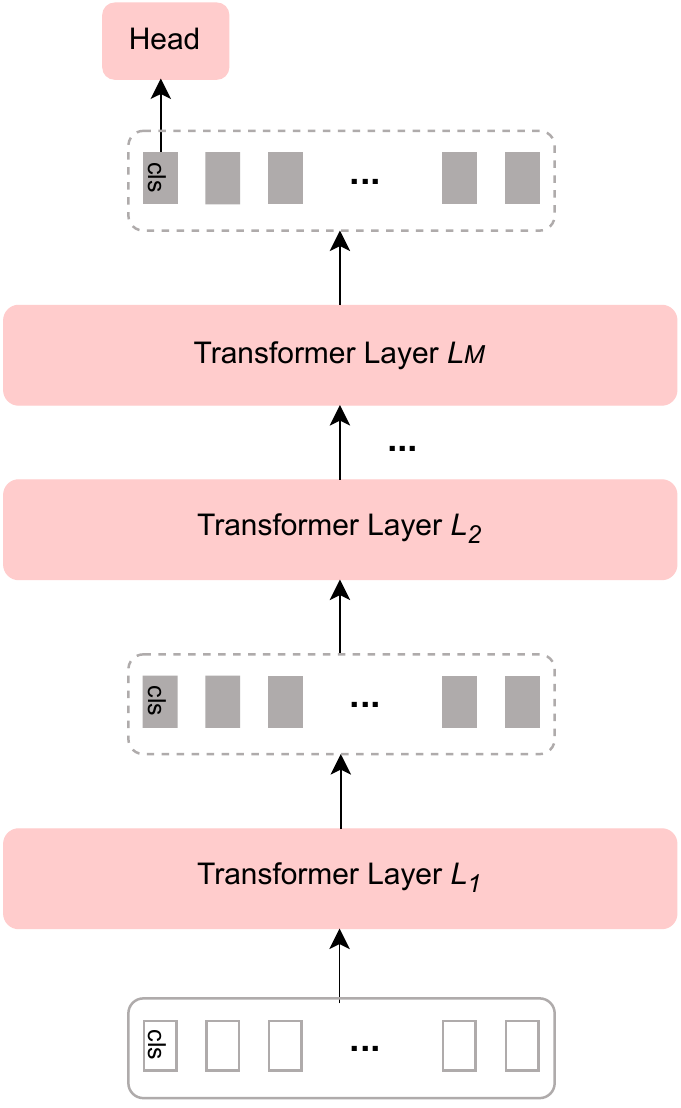}
    \caption{}
    \label{fig:finetune}
  \end{subfigure}
  \hfill
    \begin{subfigure}{0.193\textwidth}
    \includegraphics[width=\linewidth]{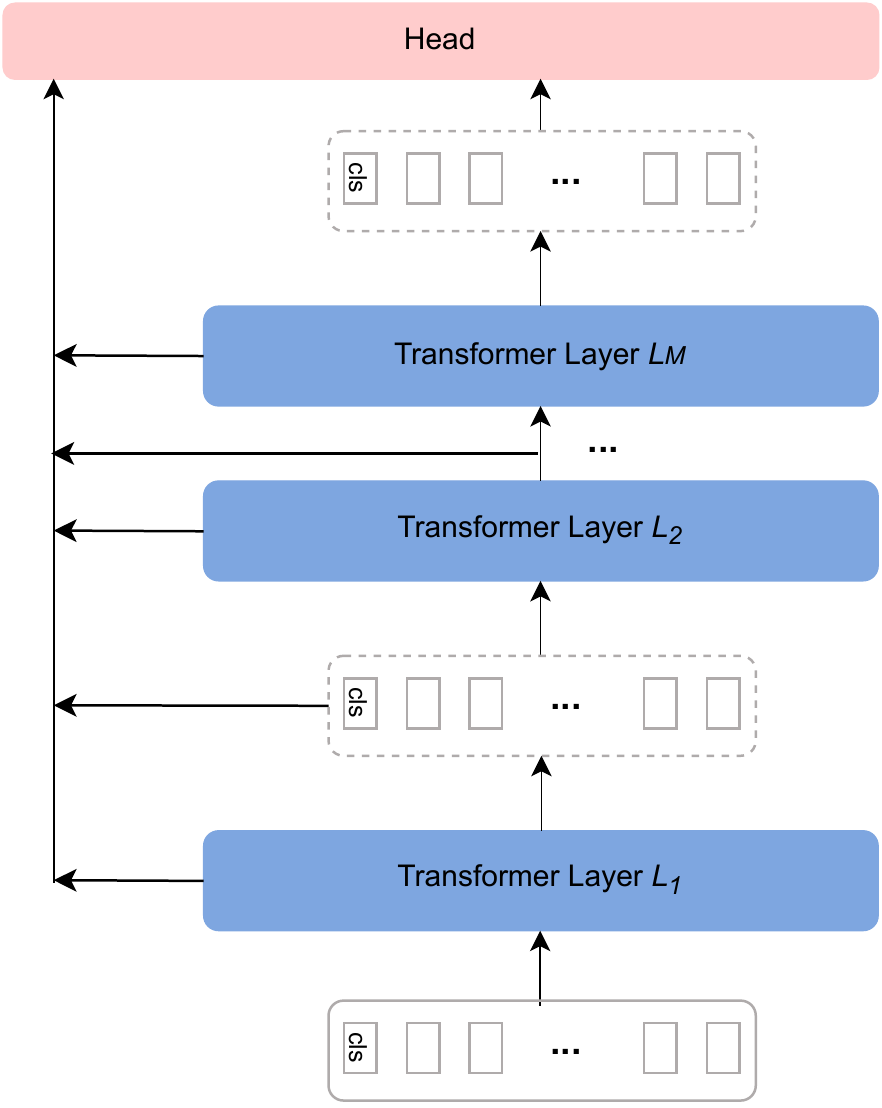}
    \caption{}
    \label{fig:h2t}
  \end{subfigure}
  \hfill
    \begin{subfigure}{0.273\textwidth}
    \includegraphics[width=\textwidth]{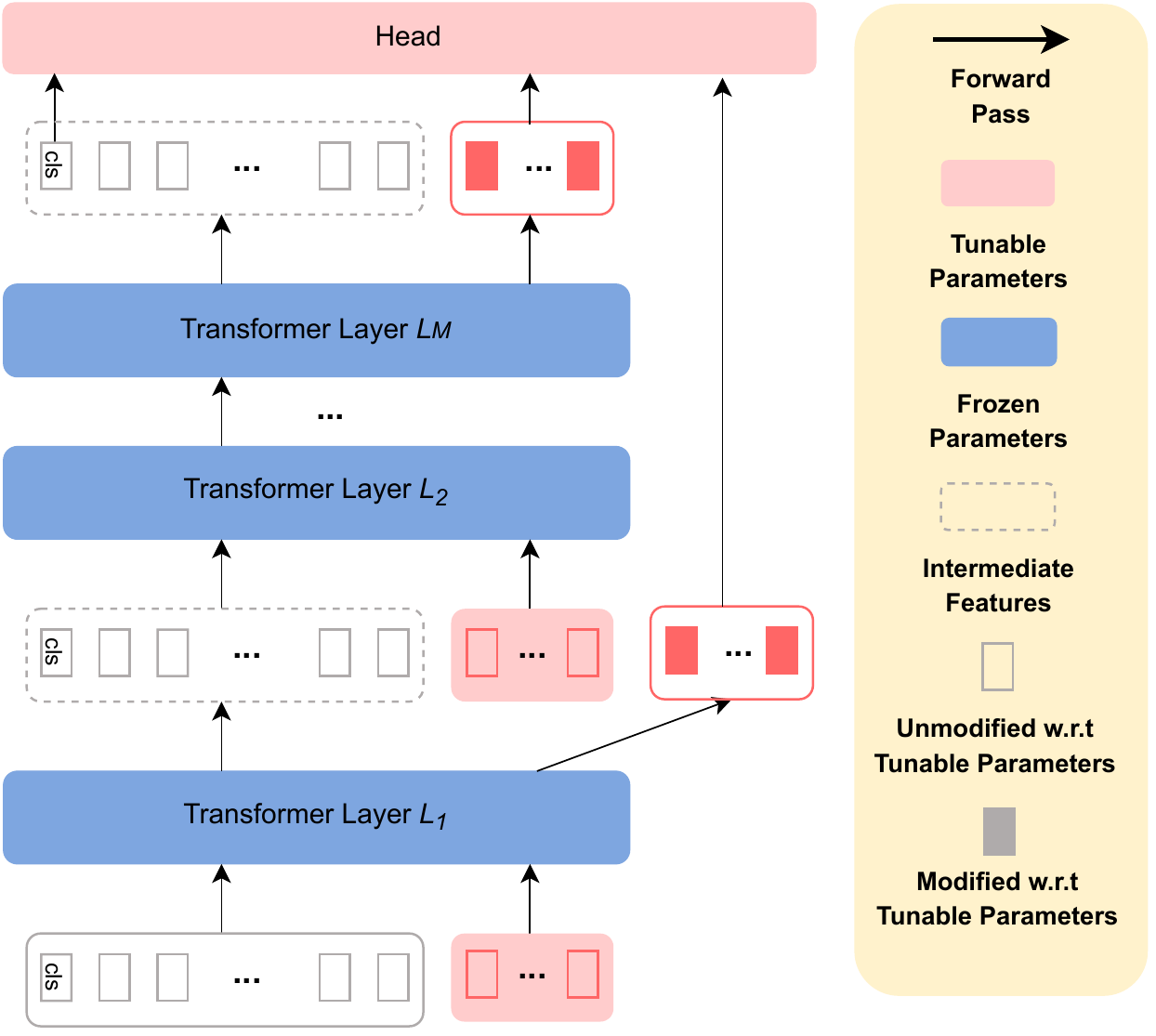}
    \caption{}
    \label{fig:vqt_compact}
  \end{subfigure}

    \caption{\textbf{Conceptual comparison between different transfer learning methods} (a) ViT Backbone. (b) Linear-Probing. (c) Fine-Tuning. (d) Head2Toe~\cite{evci2022head2toe}. (d) VQT (Ours).}
   \label{fig:compare_transfer}
\end{figure*}

\section*{\LARGE Appendix}

We provide details omitted in the main paper. 

\begin{itemize}
    \item \autoref{suppl-sec:concept_compare}: a conceptual comparison of VQT with other transfer learning methods. 
    \item \autoref{suppl-sec:exp_details}: additional experiment details (cf. \autoref{exp} of the main paper).
    \item \autoref{suppl-sec:additional_exp}: additional experiment results and analyses (cf. \autoref{exp} of the main paper). 
    \item \autoref{suppl-sec:additional_discussions}: additional discussions. 
\end{itemize}

\section{Conceptual Comparison between VQT and Transfer Learning Methods} \label{suppl-sec:concept_compare}

% \zheda{The new figure 1 and comparison of VQT with transfer learning methods}
% \zheda{TODO add color to text }
We demonstrate the conceptual difference between various transfer learning methods in \autoref{fig:compare_transfer}. \autoref{fig:vit} shows the pre-trained ViT backbone. Linear-probing (\autoref{fig:linear}) only updates the prediction head  and keeps the rest of the backbone unchanged, while fine-tuning (\autoref{fig:finetune}) updates the whole backbone. \HT (\autoref{fig:h2t}) takes intermediate outputs of blocks within each Transformer layer for predictions. In contrast, our VQT (\autoref{fig:vqt_compact}) leverages the summarized intermediate features of each Transformer layer for final predictions.

\section{Additional Experiment Details} \label{suppl-sec:exp_details}

\subsection{Pre-training Setups}

In \autoref{experiment: effectiveness} and  \autoref{experiment: compatible} of the main paper, we conduct our experiments on three types of pre-trained backbones, Supervised~\cite{dosovitskiy2021an}, MAE~\cite{he2022masked}, and CLIP~\cite{radford2021learning}. We briefly review these pre-training methods in the following.

\paragraph{Supervised.}
Given a pre-training data set $\sD_\text{pre-train} = \{(\mI_i, y_i)\}_{i=0}^N$, where $\mI_i$ is an image and $y_i \in [C]$ is the annotated class label, we aim to train a network that classifies images into $C$ classes. The network consists of a backbone network $f$ to extract features and a linear classifier $h$ to predict class labels. 
% Specifically, we first extract features of the image $\vx_i$ by a backbone network $f$. Then, the extracted features are processed via a linear classifier $h$ to predict the class label $y_i$. 
Specifically, let $\vp_i = h(f(\mI_i)) \in \R^C$ be the output of the whole network. Each element of $\vp_i$ represents the score of $\mI_i$ being classified to each of the $C$ classes. 
We apply the standard cross-entropy loss to maximize the score of classifying $\mI_i$ to be the class $y_i$. 
After pre-training, we discard the classifier $h$ and only keep the pre-trained backbone $f$ for learning downstream tasks. 

In our experiments, we use ViT-B/16 as the backbone architecture. 
In \autoref{experiment: effectiveness}, we use the ImageNet-1K pre-trained backbone following \HT~\cite{evci2022head2toe}. 
In \autoref{experiment: compatible}, we use the ImageNet-21K pre-trained backbone following VPT~\cite{jia2022vpt}. To save the pre-training time, we use the checkpoints of these backbones released on the official GitHub page of Vision Transformer~\cite{dosovitskiy2021an}\footnote{\url{https://github.com/google-research/vision_transformer}}.

\paragraph{MAE.}
The learning objective of MAE is to reconstruct an image from its partial observation.
Specifically, we divide an input image $\mI$ into $N$ fixed-sized non-overlapping patches $\{\mI^{(n)}\}_{n=1}^{N}$ following ViT~\cite{dosovitskiy2021an}.
Then, we randomly mask $K\%$ of the patches. 
Let $\sU$ be the set of indices of the unmasked patches and $|\sU| = (1-K\%) \times N$. The goal is to reconstruct the masked patches using the unmasked ones $\{\mI^{(i)} | i \in \sU\}$. 
To achieve this, we first process the unmasked patches by a ViT encoder $f$ to generate the output $\mZ_M = [\vx_{M}^{(i_1)}, \cdots, \vx_{M}^{(i_{|\sU|})}]$, where $i_1, \cdots, i_{|\sU|} \in \sU$. 
Then, we expand $\mZ_M$ to have $N$ tokens by filling $K\% \times N$ mask tokens $\vx^\text{(Mask)}$ into the positions of the masked patches to generate $\Tilde{\mZ}_M$. 
The mask token $\vx^\text{(Mask)}$ is a learnable parameter and indicates the missing patches to be predicted. 
Finally, we use a decoder $h$ to generate the reconstructed image $\Tilde{\mI} = h(\Tilde{\mZ}_M)$.
The whole encoder-decoder network is trained by comparing $\Tilde{\mI}$ with $\mI$ by using the mean squared error (MSE). Similar to BERT~\cite{kenton2019bert}, the loss is computed only on the masked patches. 
After pre-training, we discard the decoder $h$ and only keep the encoder $f$ as the backbone. 
In \autoref{experiment: compatible}, we use the ViT-B/16 backbone released in the official MAE~\cite{he2022masked} GitHub page\footnote{\url{https://github.com/facebookresearch/mae}}.

\paragraph{CLIP.}

CLIP leverages text captions of images as supervisions for pre-training a visual model. 
The learning objective is to predict which text caption is paired with which image within a batch. 
Specifically, given a batch of image-caption pairs $\{(\mI_i, \mat{C}_i)\}_{i=1}^B$, where $\mI_i$ is an image and $\mat{C}_i$ is the caption, CLIP uses a image encoder $f$ and a text encoder $h$ to map $\mI_i$ and $\mat{C}_i$ into a multi-modal embedding space, respectively. 
Let $\mZ_I = [f(\mI_1), \cdots, f(\mI_B)] \in \R^{D\times B}$ and $\mZ_C = [h(\mat{C}_1), \cdots, h(\mat{C}_B)] \in \R^{D\times B}$ be the output image and text features. 
We then compute pair-wise similarity between the columns of $\mZ_I$ and $\mZ_C$, resulting in $\mS = \mZ_I^T \mZ_C \in \R^{B\times B}$.
The diagonal elements in $\mS$ are the scores for the correct image-caption pairings while the rest elements are incorrect pairings.
CLIP minimizes the cross-entropy losses computed on the rows and the columns of $\mS$ to learn $f$ and $h$. 
After pre-training, we discard $h$ and keep the vision encoder $f$ as the pre-trained backbone. 
In \autoref{experiment: compatible}, we use the ViT-B/16 backbone released in the official CLIP~\cite{radford2021learning} GitHub page\footnote{\url{https://github.com/openai/CLIP}}.

\subsection{Feature Selection via Group Lasso} \label{suppl-sec:feat_select}

We provide more details for feature selection based on group lasso as mentioned in~\autoref{approach: vqt} of the main paper.
In VQT, we concatenate the newly summarized features $\{\mZ_{m+1}^\prime\}_{m=0}^{M-1}$ with the final ``CLS" token $\vx_M^\text{(Class)}$ for linear probing. 
Let $\mat{H}_\text{all} \in \R^{MDT+D}$ be the concatenated features and $\mat{W}_\text{all} \in \R^{(MDT+D)\times C}$ be the weights of the linear classifier, where $C$ is the number of classes. 
After we learn the additional query tokens in VQT, we can freeze them and optionally employ group lasso to reduce the dimensionality of $\mat{H}_\text{all}$. 
Specifically, we follow \HT~\cite{evci2022head2toe} to first train the linear classification head with the group-lasso regularization $|\mat{W}_\text{all}|_{2,1}$, which encourages the $\ell_2$ norm of the rows of $\mat{W}_\text{all}$ to be sparse. 
Then, the importance score of the $i$-th feature in $\mat{H}_\text{all}$ is computed as the $\ell_2$ norm of the $i$-th row of $\mat{W}_\text{all}$. 
Finally, we select a fraction $F$ of the features $\mat{H}_\text{all}$ with the largest importance scores and train a new linear head with the selected features. 

\subsection{Parameter Efficiency in Section 4.2} \label{suppl-sec:main_compare_h2t}

We provide the number of parameters for the transfer learning methods compared in~\autoref{experiment: effectiveness} of the main paper.

\paragraph{Linear-probing and full fine-tuning.}

For each task, linear probing only trains the prediction head and keeps the whole pre-trained backbone unchanged.
Therefore, we only need to maintain one copy of the backbone that can be shared across all the downstream tasks.
Contrastingly, full fine-tuning updates the whole network, including the backbone and the head, for each task. 
After training, each task needs to individually store its own fine-tuned backbone, thereby requiring more parameters.

% We further provide the total number of parameters needed for all the 19 tasks. Fine-tuning and Scratch use $19.01\times$ of a ViT-B/16 backbone parameters because they need to maintain an individual model for each task without parameter sharing. Linear-probing only uses $1.01\times$ because we only add a linear head that uses the final ``CLS" features for each task. Finally, \HT and VQT use $1.20\times$ and $1.22\times$ of the parameters, respectively.  

\paragraph{VQT vs. \HT.} 
In~\autoref{experiment: effectiveness} of the main paper, We compare VQT with \HT by matching the number of tunable parameters for each task in VTAB-1k. We provide details for this setup. More comparisons of VQT and \HT can be found in \autoref{suppl-sec:more_comp_to_h2t}. 

\HT~\cite{evci2022head2toe} takes intermediate features from multiple distinct steps inside the pre-trained ViT backbone. 
% Therefore, \HT needs a larger head than linear-probing does because it increases the feature dimension. 
For the features from each step, \HT chooses a window size and a stride to perform average pooling to reduce the dimensionality. 
After concatenating the pooled intermediate features, \HT further decides the fraction $F$ for feature selection. 
In \HT, the pooling window size, pooling stride and $F$ are hyper-parameters picked based on validation accuracy. 

% The pooling window size, pooling stride and the fraction $F$ are hyper-parameters picked based on the validation accuracy for each task in VTAB-1k. 
For fair comparisons of VQT and \HT~\cite{evci2022head2toe}, we match their numbers of tunable parameters used in different tasks. 
As both VQT and \HT leverage intermediate features while keeping the backbone unchanged, the tunable parameters mainly reside in the final linear head. 
We focus on matching the feature dimension input to the classifier. 
First, we divide the 19 tasks in VTAB-1k into three groups; each group corresponds to a pair of pooling window sizes and strides that \HT uses to generate the features \emph{before} feature selection. 
Specifically, in the three groups, \HT generates 68K-, 815K-, and 1.8M-dimensional features, respectively. 
Next, for simplicity, we set $F = 0.1$, which is the maximal $F$ in \HT's hyper-parameter search grid, for \HT in all the 19 tasks. 
These results are obtained using the official \HT released code\footnote{\url{https://github.com/google-research/head2toe}}.
For VQT, we choose $T$ and $F$ to match the final feature dimensions (after feature selection) used in \HT. 
Specifically, we use $T=1$ and $F=0.7$, $T=10$ and $F=1.0$, and $T=20$ and $F=1.0$ for the three task groups, respectively.

\paragraph{Comparison on the number of parameters.}
We summarize the number of parameters needed for all the 19 VTAB-1k tasks in \autoref{table:number_of_params}.
As linear-probing shares the backbone among tasks and only adds linear heads that take the final ``CLS" token, it only requires $1.01\times$ of the ViT-B/16 backbone parameters. 
By contrast, fine-tuning consumes $19.01\times$ of the ViT-B/16 because each task maintains its own fine-tuned backbone. 
Compared to linear probing, VQT and \HT use larger feature dimensions for prediction, thereby increasing the number of parameters in the final linear heads.
Even so, VQT and \HT are still parameter-efficient and need only $1.22\times$ and $1.20\times$ of the backbone parameters to learn 19 tasks.

\begin{table}[t]
    \centering
    \begin{tabular}{c|c}
        \toprule
        Methods & \makecell{Total  \\\# of parameters} \\
        \midrule
        Scratch             & $19.01\times$ \\
        Linear-probing      & $1.01\times$ \\
        Fine-tuning         & $19.01\times$ \\
        \HT                 & $1.20\times$ \\
        \textbf{VQT (Our)}  & $1.22\times$ \\
        \bottomrule
    \end{tabular}
    \vskip-5pt
    \caption{Total numbers of parameters needed for all the 19 tasks, for the methods compared in~\autoref{table:h2t}. Each number represents how many times of one ViT-B/16 backbone's parameters (86M) are needed.}
    \vskip-10pt
    \label{table:number_of_params}
\end{table}

% - Table 1: include a column to show the number of parameters

% - Section 4.3 the number of parameters

% === Compare to Heat-2-Toe === 
% \begin{itemize}
%     \item same parameters (together with \# of parameters)
% \end{itemize}

% \begin{figure*}[tb]
%     \centering
%     \includegraphics[width=\linewidth]{CVPR 23/figures/detailed_robust.pdf}
%     \caption{\zheda{some plots's blue and orange are wrong, double check all. }}
%     \label{fig:robust_fraction}
% \end{figure*}

\subsection{Additional Training Details} 
\label{suppl_sub_train_detail}
We provide training details for VQT, VPT~\cite{jia2022vpt} and AdaptFormer~\cite{chen2022adaptformer} used in \autoref{exp}. 

For each task in VTAB-1k, we perform an 80/20 split on the 1K training images to form a training/validation set for hyper-parameter searching. After we pick the best hyper-parameters that yield the best validation accuracy, we use them for training on the original 1K training images. Finally, we report the accuracy on the testing set. 

For training VQT with the ImageNet-1K backbone, we set $T$ to match the number of tunable parameters with \HT as described in \autoref{suppl-sec:feat_select}. For training VQT with the MAE backbone (results reported in \autoref{tab:mae}), we simply set $T=1$ for all tasks in VTAB-1k.
We then perform a hyper-parameter search to select the best learning rate from $\{1.0, 0.5, 0.25, 0.1, 0.05\}$ and the best weight decay from $\{0.01, 0.001, 0.0001, 0.0\}$. 
We use the Adam optimizer to train VQT for 100 epochs, and the learning rate follows the cosine schedule. 

For training VPT with the ImageNet-21K and the MAE backbones, we use each task's best number of prompt tokens, which are released in VPT's GitHub page\footnote{\url{https://github.com/KMnP/vpt}}. 
Since the VPT paper does not use the CLIP backbone, we simply set the number of tokens to $1$ for all tasks in this case. We conduct a hyper-parameter search to pick the best learning rate from $\{1.0, 0.5, 0.25, 0.1, 0.05\}$ and the best weight decay from $\{0.01, 0.001, 0.0001, 0.0\}$. We train VPT using the Adam optimizer for 100 epochs with the cosine learning rate schedule. 

Finally, when training AdaptFormer on all backbones, we use the bottleneck dimension $\hat{d} = 64$ and the scaling factor $s = 0.1$ following~\cite{chen2022adaptformer}. We similarly search the best learning rate from $\{1.0, 0.5, 0.25, 0.1, 0.05\}$ and the best weight decay from $\{0.01, 0.001, 0.0001, 0.0\}$ using the validation set. The AdaptFormer is trained with the Adam optimizer for 100 epochs, and the learning rate decay follows the cosine schedule.

% Datasets. Hyerparameter search. 

\section{Additional Experiments and Analyses} \label{suppl-sec:additional_exp}

\subsection{More Comparison with \textbf{\HT}} 
\label{suppl-sec:more_comp_to_h2t}

\begin{table*}[t!]
\normalsize
\resizebox{\textwidth}{!}{
\begin{tabular}{l|ccccccc:c|cccc:c|cccccccc:c|c}
\toprule
& \multicolumn{8}{c|}{Natural}&\multicolumn{5}{c|}{Specialized}&\multicolumn{9}{c}{Structured}&\\
Method& {\rotatebox[origin=l]{90}{CIFAR-100}}&{\rotatebox[origin=l]{90}{Caltech101}}&{\rotatebox[origin=l]{90}{DTD}}&{\rotatebox[origin=l]{90}{Flowers102}}&{\rotatebox[origin=l]{90}{Pets}}&{\rotatebox[origin=l]{90}{SVHN}}&{\rotatebox[origin=l]{90}{Sun397}}&{\rotatebox[origin=l]{90}{Mean}}&{\rotatebox[origin=l]{90}{Camelyon}}&{\rotatebox[origin=l]{90}{EuroSAT}}&{\rotatebox[origin=l]{90}{Resisc45}}&{\rotatebox[origin=l]{90}{Retinopathy}}&{\rotatebox[origin=l]{90}{Mean}}&{\rotatebox[origin=l]{90}{Clevr-Count}}&{\rotatebox[origin=l]{90}{Clevr-Dist}}&{\rotatebox[origin=l]{90}{DMLab}}&{\rotatebox[origin=l]{90}{KITTI-Dist}}&{\rotatebox[origin=l]{90}{dSpr-Loc}}&{\rotatebox[origin=l]{90}{dSpr-Ori}}&{\rotatebox[origin=l]{90}{sNORB-Azim}}&{\rotatebox[origin=l]{90}{sNORB-Elev}}&{\rotatebox[origin=l]{90}{Mean}}&{\rotatebox[origin=l]{90}{Overall Mean}}\\
\midrule
\HT\ & 58.2 &87.3 &64.5 &85.9 &85.4 &82.9 &35.1 &71.3 &81.2 &95.0 &79.9 &74.1 &82.6 &49.3 &\textbf{58.4} &41.6 &64.4 &53.3 &32.9 &\textbf{33.5} &\textbf{39.4} &46.6 &63.3 \\
\textbf{VQT (Ours)} & \textbf{58.5} &\textbf{89.5} &\textbf{66.7} &\textbf{89.9} &\textbf{88.8} &\textbf{79.7} &35.1 &\textbf{72.6} &\textbf{82.4} &\textbf{96.2} &\textbf{84.4} &\textbf{74.8} &\textbf{84.5} &\textbf{50.5} &57.1 &\textbf{42.7} &\textbf{77.9} &\textbf{69.2} &\textbf{43.6} &24.1 &32.0 &\textbf{49.6} &\textbf{65.4} \\
\bottomrule
\end{tabular}}
\caption{\HT and VQT's test accuracies on the VTAB-1k benchmark with ViT-B/16 pre-trained on ImageNet-1K. In this comparison, we do not set parameter constraints and use the validation set to choose the best feature dimension based on accuracy. "Mean" denotes the average accuracy for each category and "Overall Mean" shows the average accuracy over 19 tasks.} \label{table:h2t_noconstraint}
\end{table*}

In~\autoref{table:h2t} of the main paper, we have compared VQT with \HT under the constraint of using a similar number of tunable parameters, as mentioned in \autoref{suppl-sec:main_compare_h2t}. 
% We provide more detailed comparison between the two methods. 
%\paragraph{Without the constraint on parameter size.} 
To further evaluate the limit of VQT, we drop this constraint 
% of matching the parameter sizes of VQT and \HT 
and allow both VQT and \HT to select the best feature dimensions based on accuracy. 
Specifically, we pick the best feature fraction $F$ for VQT using the validation set and compare it with the \emph{best} \HT results, which are also obtained by selecting the best feature dimension via hyper-parameter tuning, reported in their paper~\cite{evci2022head2toe}. 
\autoref{table:h2t_noconstraint} shows the results of \HT and VQT without the parameter constraint. 
VQT still significantly outperforms \HT on 15 out of 19 tasks across the Natural, Specialized and Structured categories of VTAB-1k, demonstrating the effectiveness of the summarized intermediate features in VQT. We also compare \HT and VQT on different pre-trained setups. As shown in \autoref{tab:h2t_vqt_detail_backbone},  VQT consistently outperforms \HT on supervised,  self-supervised (MAE) and image-language (CLIP) pre-trained backbones. We used the best hyper-parameters from the \HT paper for the ImageNet-1K backbone, and we only performed the learning rate and weight decay hyper-parameters search for the MAE and CLIP model. We match the number of tunable parameters in VQT and \HT for fair comparisons.

\begin{table}[t]
    \footnotesize
    \centering
    \begin{tabular}{cccc|c}
        \toprule
        Methods & Natural & Specialized & Structured& Mean \\
        \midrule
        \multicolumn{5}{c}{ImageNet-1K }\\
        \midrule
        H2T& 68.9&82.9&46.3&62.3\\
        % VPT (P=1)&75.5&81.2&45.2&63.9\\
        % AdaptFormer&73.4&80.1&47.3&63.8\\
        \textbf{VQT}&\textbf {72.7}&\textbf{84.5}&\textbf{49.3}&\textbf{65.3}\\
        % \midrule
        % \multicolumn{5}{c}{ImageNet-21K }\\
        % \midrule
        % HEAD2TOE&69.8&82.2&30.3&56.0\\
        % % VPT&79.1&84.6&56.0&69.9\\
        % % AdaptFormer&80.1&82.3&50.3&68.0\\
        % {VQT}&{74.7}&{83.5}&{46.4}&64.4\\
        \midrule
        \multicolumn{5}{c}{MAE}\\
        \midrule
        H2T&55.6&80.3&44.4& 55.7\\
        % VPT&63.5&79.1&48.6&60.5\\
        % AdaptFormer&68.7&81.3&58.3&67.0\\
        \textbf{VQT}&\textbf{66.0}&\textbf{82.9}&\textbf{53.5}&\textbf{63.9}\\
        \midrule
        \multicolumn{5}{c}{CLIP}\\
        \midrule
        H2T&69.3&82.0&33.8&57.0\\
        % VPT&63.5&79.1&48.6&60.5\\
        % AdaptFormer&68.7&81.3&58.3&67.0\\
        \textbf{VQT}&\textbf{77.7}&\textbf{83.7}&\textbf{51.3}&\textbf{67.9}\\
\bottomrule

    \end{tabular}
    \vskip-5pt
    \caption{Performance comparison between HEAD2TOE (H2T) and VQT on various backbones. }
    \vskip-10pt
    \label{tab:h2t_vqt_detail_backbone}
\end{table}

\begin{figure}[tb]
    \centering
    \includegraphics[width=\linewidth]{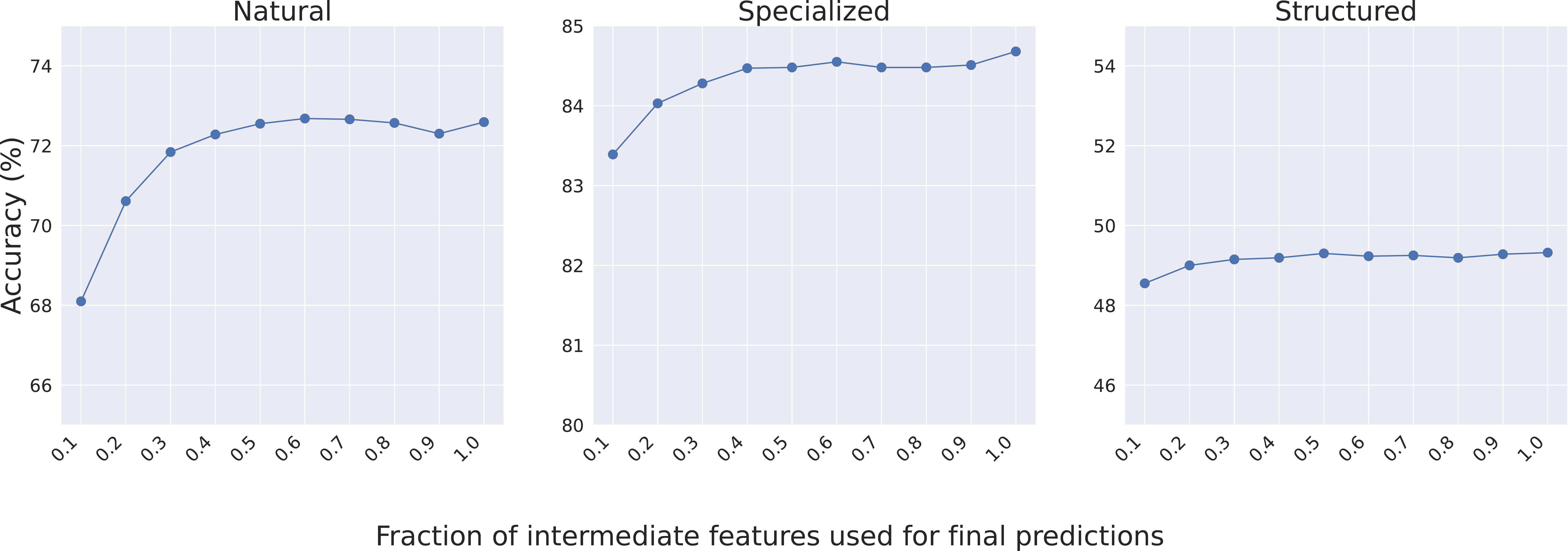}
    \caption{Average accuracy on VTAB-1k using different fractions of intermediate features for VQT.}
    \label{fig:robust_fraction}
\end{figure}

\subsection{Robustness to Different Feature Fractions} 
We study the robustness of VQT using different fractions of the intermediate features for prediction. Given a fraction $F$, we follow the strategy mentioned in~\autoref{suppl-sec:feat_select} to select the best features. 
As shown in~\autoref{fig:robust_fraction}, VQT is able to maintain its accuracy, with less than $1\%$ drop, even when we discard $60\%$ of the features.
On the Structured category in VTAB-1k, we can even drop up to $90\%$ of the features for VQT without largely degrading its performance. 
These results reveal the potential of further compressing VQT to reduce more parameters.

% \begin{itemize}
%     \item no constraints 
%     \item stability / robustness
% \end{itemize}

\subsection{Different Vision Transformer Backbones} 
\begin{figure}[tb]
    \centering
    \includegraphics[width=\linewidth]{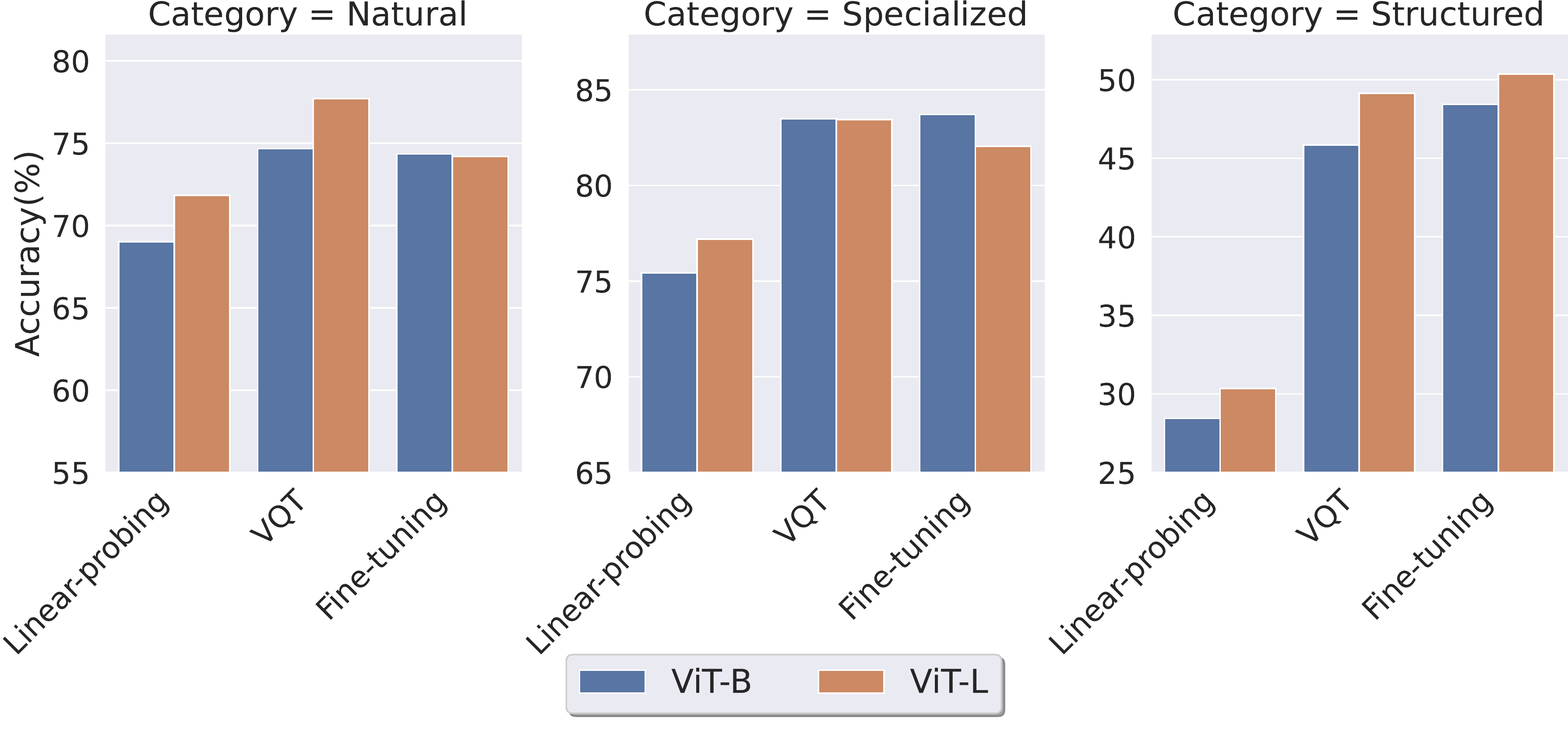}
    \caption{Performance comparison between \emph{linear-probing}, \emph{fine-tuning} and VQT on ViT-\textbf{B}ase (86M parameters) and ViT-\textbf{L}arge (307M parameters) pretrained on ImageNet-21K}
    \label{fig:backbone_scale}
\end{figure}

\autoref{fig:backbone_scale} shows the performance comparison between \emph{linear-probing}, \emph{fine-tuning} and VQT on ViT-\textbf{B}ase (86M parameters) and ViT-\textbf{L}arge (307M parameters) pretrained on ImageNet-21K. Generally speaking, all methods perform better on ViT-L than ViT-B due to higher model complexity. In the Natural and Specialized category, VQT has similar performance as \emph{fine-tuning} on ViT-B and outperforms \emph{fine-tuning} on ViT-L. As explained in \autoref{experiment: effectiveness}, the Natural and Specialized categories have stronger domain affinities with the source domain (ImageNet). Thus, both pre-trained backbones can generate more relevant intermediate features for similar domains. In the Structured category, \emph{fine-tuning} slightly surpasses VQT on both backbones due to the difference between the pretrained dataset and the Structured category.

\subsection{Variants of VQT} 
\label{suppl-sec:variants}
We ablate different design choices on the ViT-B pretrained on ImageNet-21K and evaluate them on the VTAB dataset.  

\begin{figure*}
\centering
  \begin{subfigure}{0.49\linewidth}
    \includegraphics[width=\linewidth]{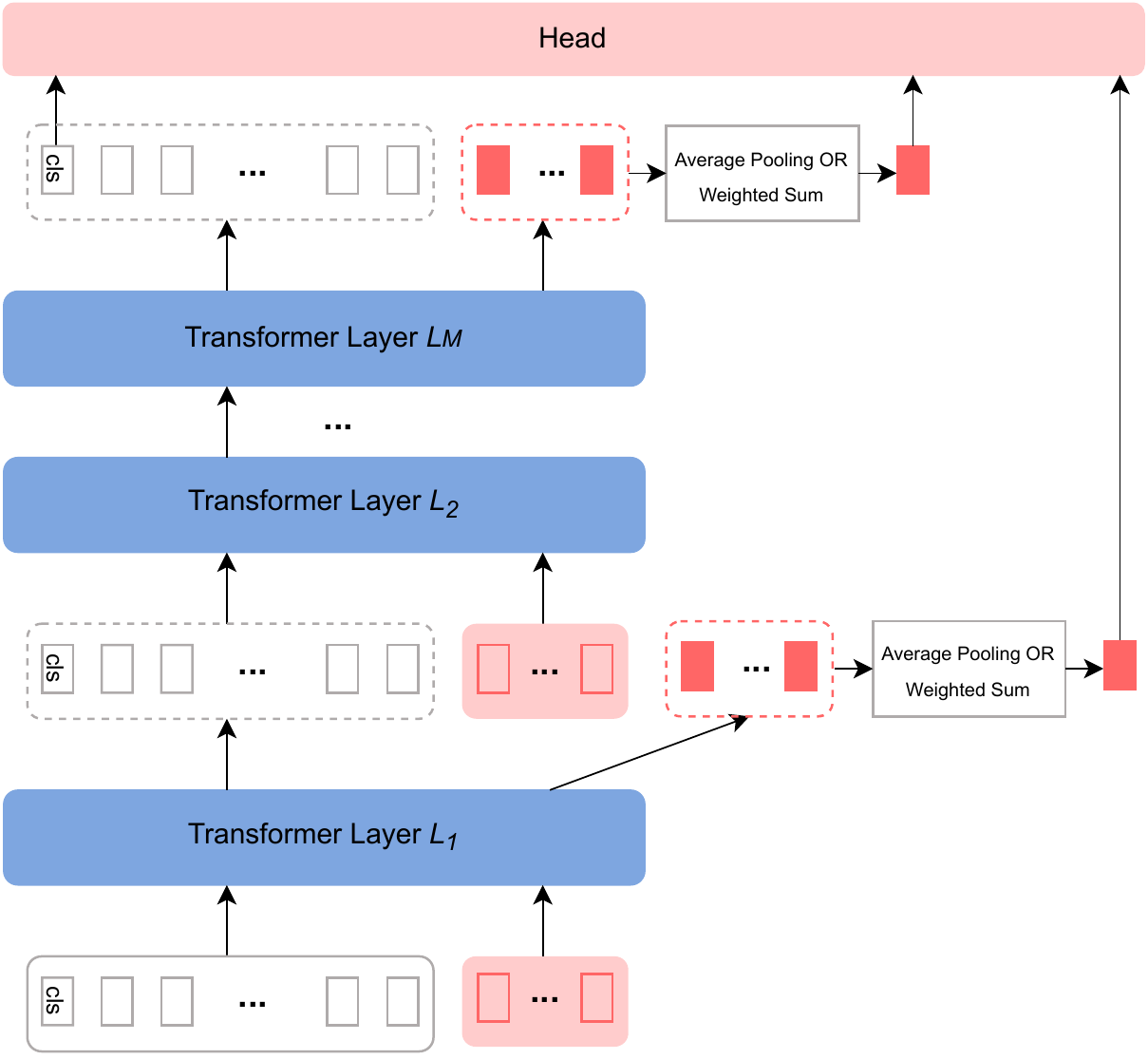}
    \caption{}
    \label{fig:within}
  \end{subfigure}
  \hspace{0.1\textwidth}
  \begin{subfigure}{0.32\linewidth}
    \includegraphics[width=\linewidth]{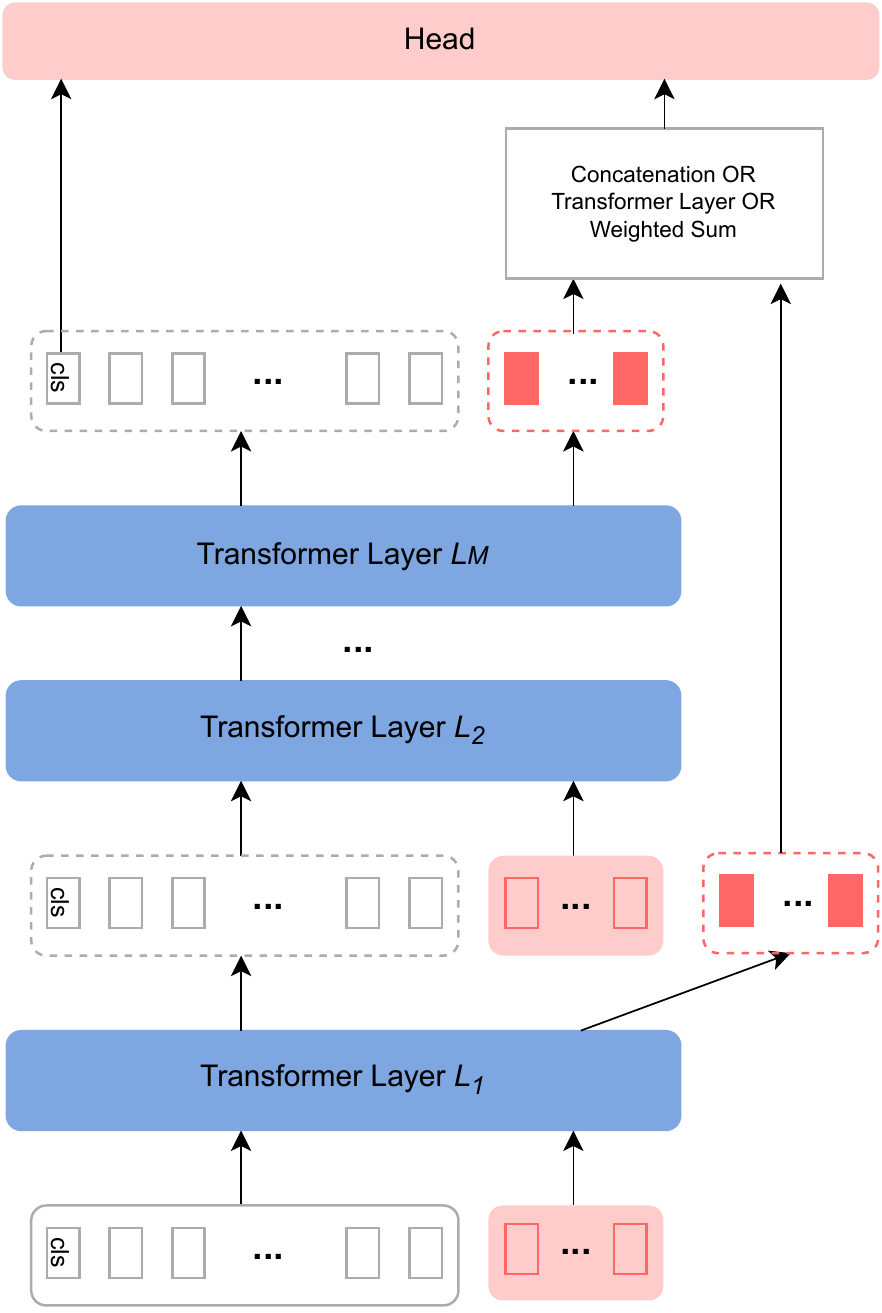}
    \caption{}
    \label{fig:across}
  \end{subfigure}
    \caption{(a) shows the within-layer aggregate methods. Multiple output query tokens within the same layer can be aggregated through average pooling or weighted sum. (b) shows the across-layer aggregation methods. Output query tokens from different layers can be aggregated through concatenation, weighted sum or another randomly initialized Transformer layer.  }
\end{figure*}

\paragraph{Summarized feature aggregation within layers.} VQT relies on each layer's summarized features (the outputs of query tokens) for predictions. Although adding a suitable number of tokens can improve the performance as shown in \autoref{fig:q_num}, we investigate if we can effectively aggregate the summarized features within a Transformer layer to reduce the dimensionality by two approaches: (1) average pooling and (2) weighted-sum, as shown in \autoref{fig:within}. Specifically, (1) we perform pooling to average T output tokens back to \emph{1} token; (2) we learn a set of weights for each layer to perform weighted-sum over T output tokens. After the aggregation step, the size of the summarized features for each layer will be changed from $\R^{D\times T}$  to $\R^{D\times 1}$. 

\begin{figure}
\centering
  \begin{subfigure}{0.49\linewidth}
    \includegraphics[width=\linewidth]{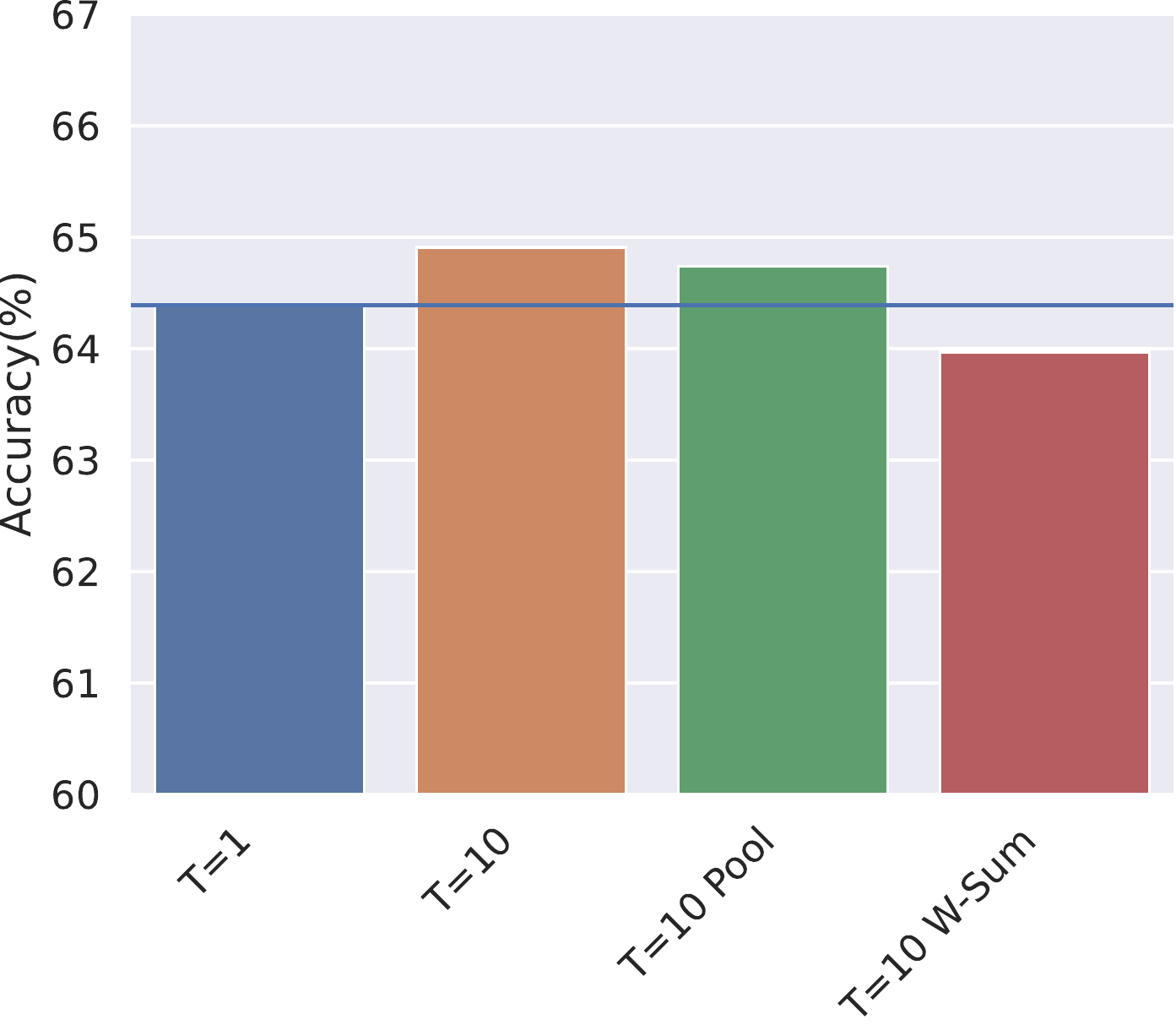}
    \caption{}
    \label{fig:within_10}
  \end{subfigure}
  \begin{subfigure}{0.49\linewidth}
    \includegraphics[width=\linewidth]{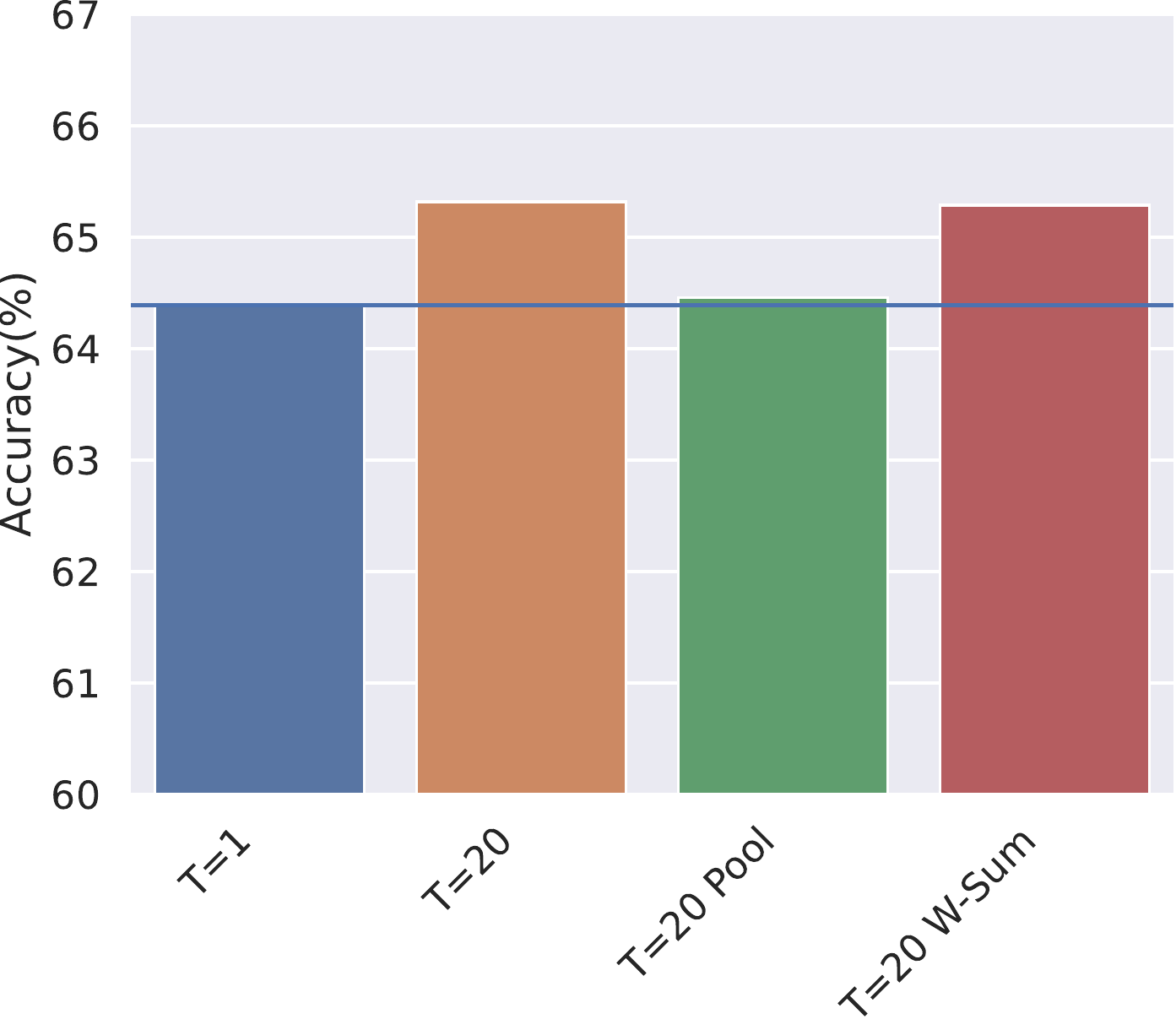}
    \caption{}
    \label{fig:within_20}
  \end{subfigure}
    \caption{Performance comparison for different within-layer aggregation methods when T=10 and T=20, where "pool" and "W-Sum" refers to average pooling and weighted sum, respectively. The blue line shows the accuracy for T=1. Note that the summarized feature dimension for T=10 (20) pool (w-Sum) is the same as the one for T=1.}
\end{figure}

\autoref{fig:within_10} and \autoref{fig:within_20} show the aggregation performance for T=10 and T=20 respectively. When we use T=10, average pooling performs similarly to T=10 and outperforms T=1 and weighted-sum. However, the trend is reversed when we use T=20; weighted-sum surpasses average pooling and T=1. To strike a balance between performance and efficiency, we suggest utilizing the validation set to choose a good within-layer aggregation method for a downstream dataset. 

\begin{figure}
    \centering
    \includegraphics[width=0.7\linewidth]{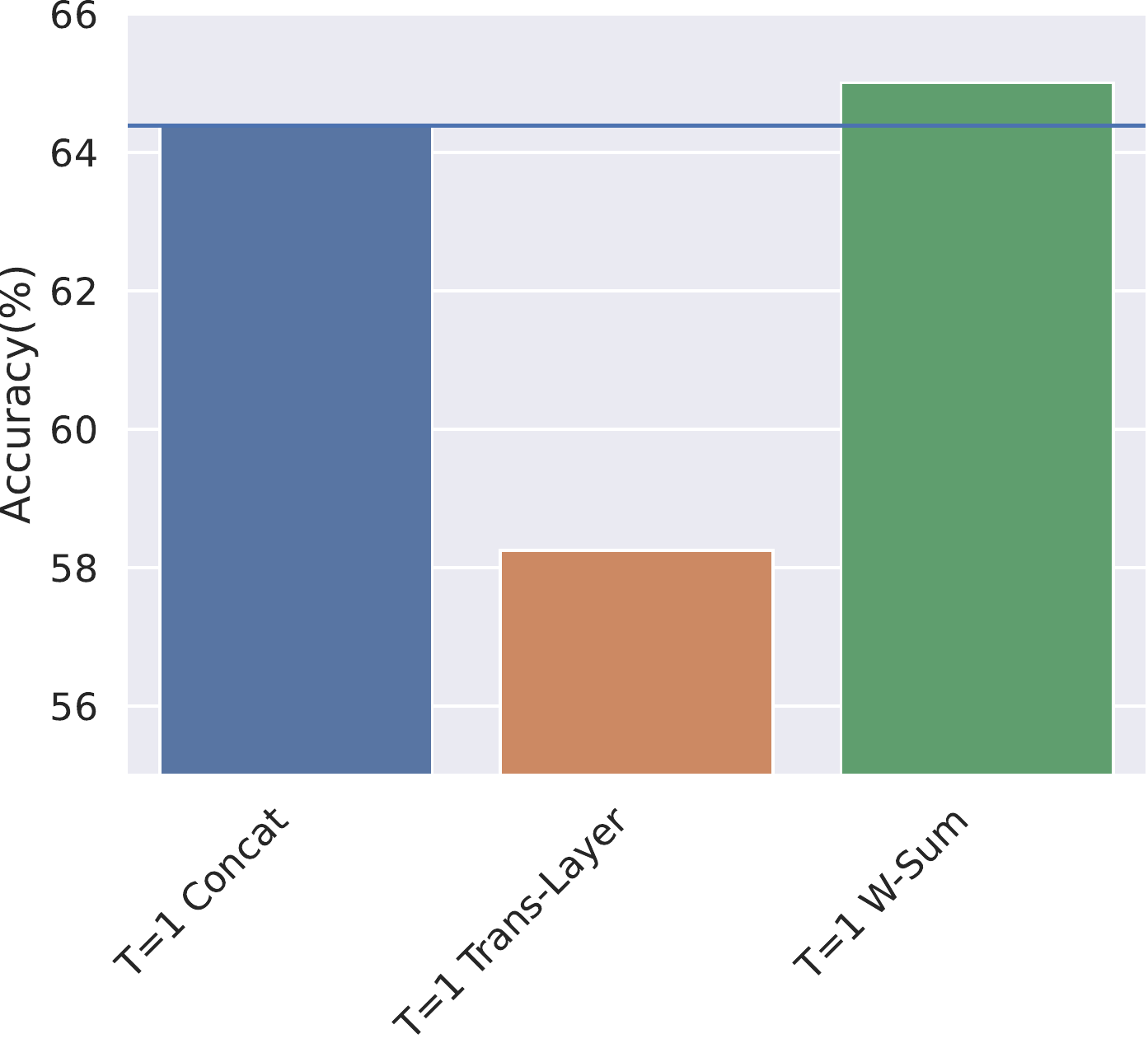}
    \caption{Performance comparison for different across-layer aggregation methods when T=1. The blue line shows the accuracy for T=1 \emph{Concat}. \emph{W-Sum} is a more efficient and effective way to aggregate summarized features across layers since it reduces the dimensionality and performs better.  }
    \label{fig:across_perf}
\end{figure}

\paragraph{Summarized feature aggregation across layers.} This section explores how to aggregate the summarized features (the outputs of query tokens) across layers. Instead of concatenation (\emph{Concat}), the default method we use in the main paper, we try feeding the summarized features from all layers to a randomly initialized Transformer layer with the CLS token from the last Transformer layer and use the output of the CLS token for prediction, dubbed \emph{Trans-Layer}. We also try to perform weighted-sum over all the summarized features, dubbed \emph{W-Sum}. When T=1, the dimension for Concat is $\R^{D\times M}$ where $M$ is the number of Transformer layers in the backbone and the dimension for \emph{Trans-Layer} and \emph{W-Sum} is $\R^{D\times 1}$. The across-layer aggregation methods are demonstrated in \autoref{fig:across}. 

As shown in \autoref{fig:across_perf}, \emph{Trans-Layer} is way behind \emph{Concat}. We hypothesize that the limited number of images per task may not be sufficient to train a randomly initialized Transformer layer. On the contrary, \emph{W-Sum} outperforms the default \emph{Concat}, which is surprising for us since the same dimension of the summarized feature in different layers may represent different information, and thus, the summarized feature from different layers may not be addable. However, based on this result, we hypothesize that the skip connection in each layer can be the cause of the addibility of summarized features from different layers. We believe studying more effective and efficient aggregation methods for the summarized features is an interesting future direction.

% suggesting that the summarized features from different layers are addable and \emph{W-Sum} is a more efficient and effective way to aggregate summarized features across layers. 

% \zheda{suprised it works skip link, keep same dimension means the same thing, it's another direction}

% \zheda{draw figures to show different methods}

% === VQT Q = 1 === 
% \begin{itemize}
%     \item concat (ours in the main paper) 
%     \item w-sum  
%     \item self-attention 
% \end{itemize}

% \subsection{Prompt-Depth Investigation} 

\subsection{t-SNE Visualization for More Datasets}
\label{suppl-sec:tsne}
\begin{figure*}
\centering
  \begin{subfigure}{0.49\textwidth}
    \includegraphics[width=\linewidth]{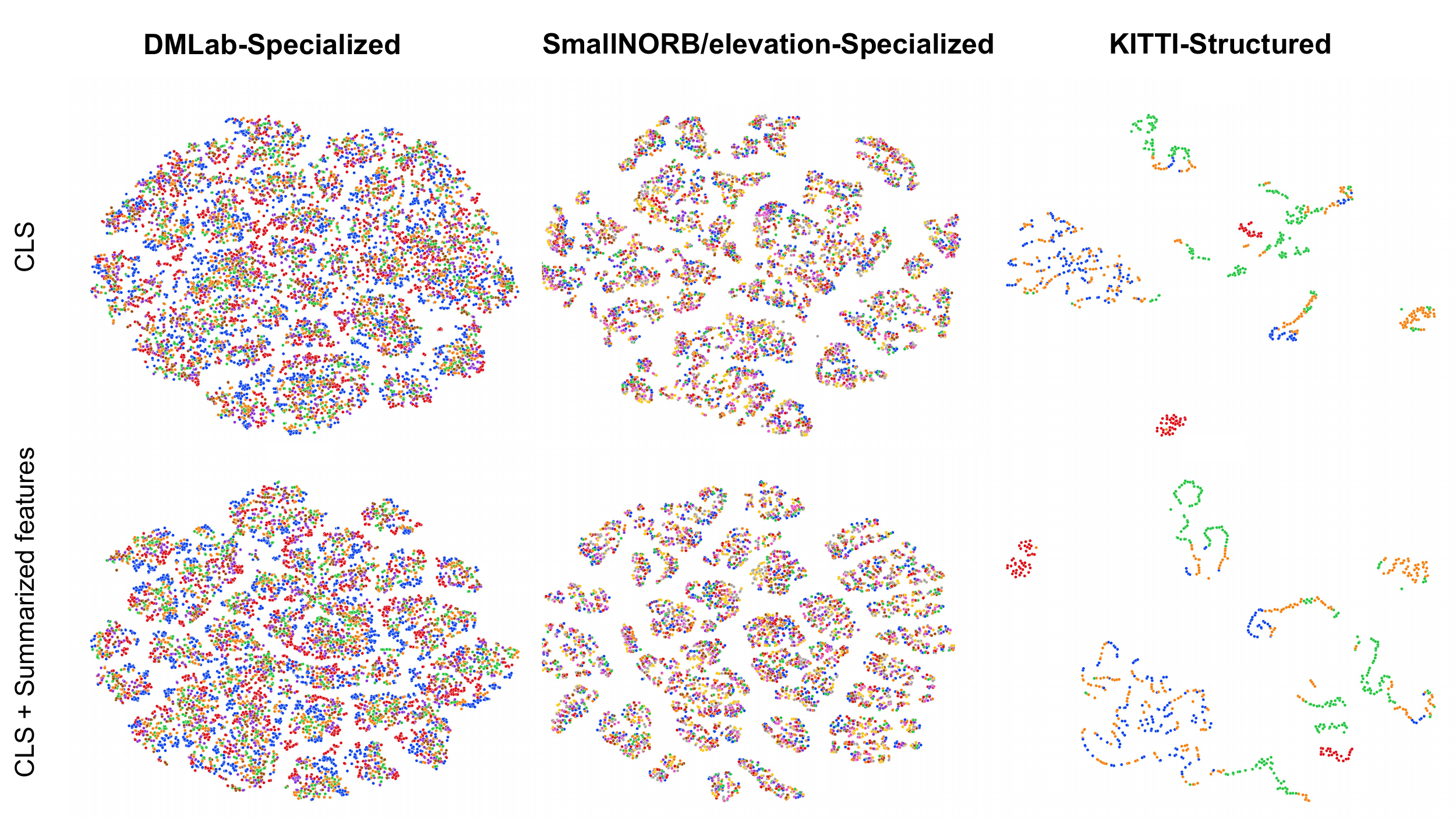}
    \caption{}
    \label{fig:tsne1}
  \end{subfigure}
  \begin{subfigure}{0.49\textwidth}
    \includegraphics[width=\linewidth]{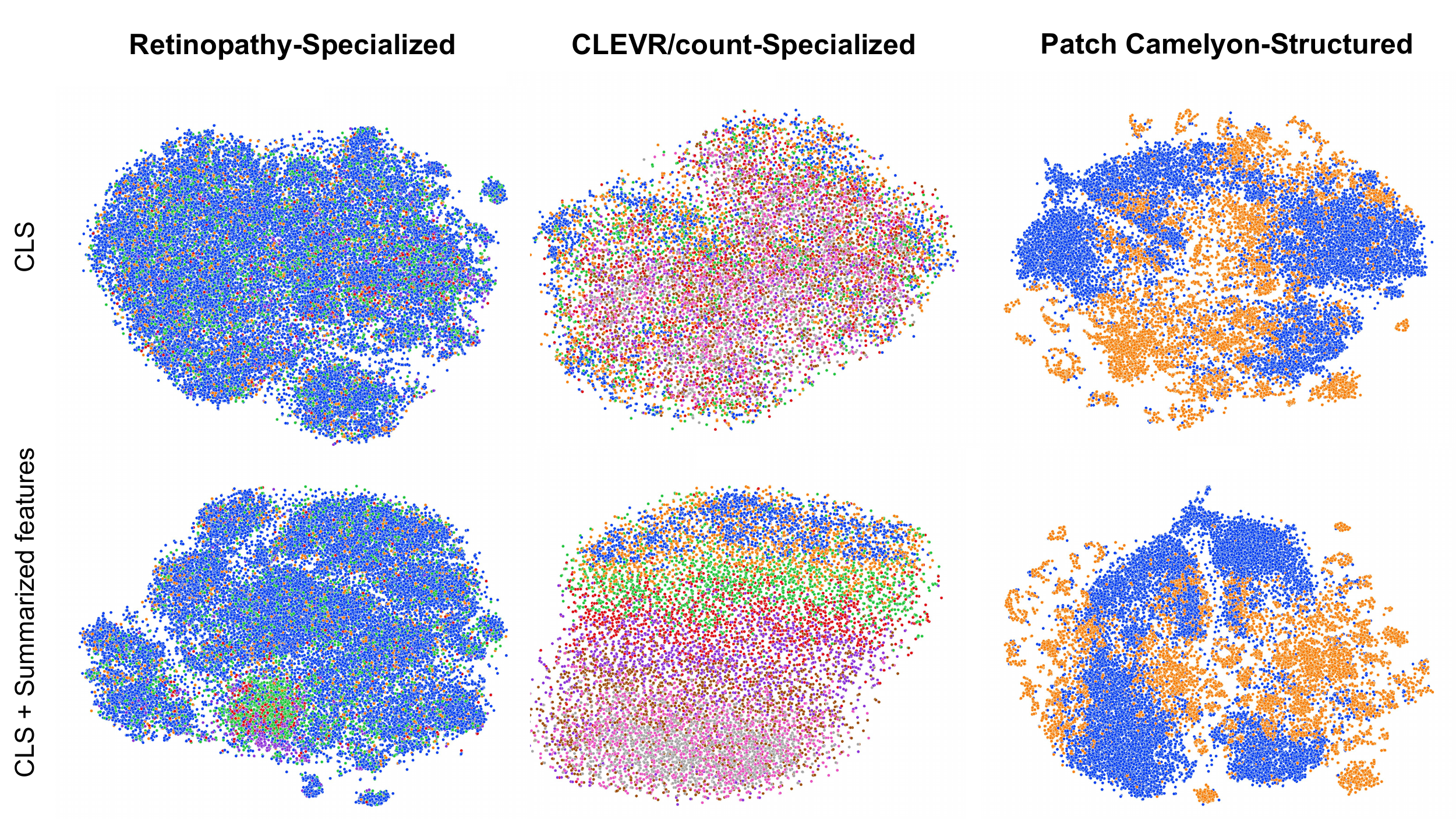}
    \caption{}
    \label{fig:tsne2}
  \end{subfigure}

    \caption{\textbf{t-SNE visualization of the CLS tokens alone (top) and CLS tokens plus our summarized features (bottom)} on more tasks from VTAB. Adding the summarized intermediate features makes the whole features more separable. We include tasks that have less or equal to 10 classes for visualization.}

   \label{fig:tsne_more}
\end{figure*}

We present t-SNE visualizations of the CLS token and our summarized features for more tasks in \autoref{fig:tsne_more}. Similar to \autoref{fig:tsne}, adding summarized features makes the whole features more separable than the CLS token alone, demonstrating the benefit of using intermediate features and the advantage of our query tokens in summarizing them.

\subsection{Results for All Tasks on Different Backbones} 
\label{suppl-sec:results_backbones}
\autoref{table:detailed_acc} shows the per-task accuracies for 19 tasks in VTAB on different ViT-B backbones, including CLIP, MAE and Supervised ImageNet-21K.

\subsection{Compatibility comparison between VQT and H2T} 
\label{suppl-sec:comptibility}
We compare the compatibility performance between HEAD2TOE and VQT with VPT and AdaptFormer (AF). For a fair comparison, we ensure that the output feature dimension is the same as the original one (D=768 in ViT) when we combine VPT and AdaptFormer with HEAD2TOE and VQT. We use the default feature selection method in the original paper for HEAD2TOE and the weighted-sum approach (see \autoref{suppl-sec:variants} for details) for VQT. \autoref{tab:h2t_vqt_compatibility} shows the results on ImageNet-1k pre-trained backbone and VQT demonstrates more competitive compatibility performance than HEAD2TOE.  

\begin{table}[t]
\small
    \centering
    \begin{tabular}{llll|l}
        \toprule
        Methods & Natural & Specialized & Structured& Mean \\
        \midrule
        % \multicolumn{4}{c}{ImageNet-1K }\\
        % \midrule
%         H2T& 68.9&82.9&46.3&62.3\\
% {VQT}& {72.7}&{84.5}&{49.3}&{65.3}\\
VPT&74.9&82.9&53.9&65.9\\
VPT+H2T&69.1&81.1&50.9&64.0\\
VPT+VQT&\textbf{76.8} (6/7)&\textbf{83.8}(2/4)&\textbf{53.4}(6/8)&\textbf{68.4}\\
AF&73.4&80.1&47.3&63.8\\
AF+H2T&69.4&82.3&51.4&64.5\\
AF+VQT&\textbf{77.0}(7/7)&\textbf{84.6}(2/4)&\textbf{53.4}(6/8)&\textbf{68.7}\\
        % \midrule
        % \multicolumn{5}{c}{ImageNet-21K }\\
        % \midrule
        
        % VPT+HEAD2TOE&80.1&84.2&55.1&70.4\\
        % VPT+VQT&78.9&83.7&54.6&69.7\\
        % AdaptFormer+HEAD2TOE&80.1&83.6&51.1&68.7\\
        % AdaptFormer+VQT&79.6&84.3&53.0&69.4\\
        % \midrule
        % \multicolumn{5}{c}{MAE}\\
        % \midrule
        % VPT+HEAD2TOE\\
        % VPT+VQT&67.9&82.7&49.7&63.4\\
        % AdaptFormer+HEAD2TOE&70.8&83.8&59.8&68.9\\
        % AdaptFormer+VQT&71.1&83.3&59.2&68.6\\

\bottomrule

    \end{tabular}
    \caption{Compatibility comparison between HEAD2TOE (H2T) and VQT on VPT and AdaptFormer (AF). The (/) represents the number of wins compared to baselines and baselines+H2T. The results are based on ImageNet-1k pre-trained
backbone. }
    \vskip-10pt
    \label{tab:h2t_vqt_compatibility}
\end{table}

\begin{table*}[t!]
\normalsize
\resizebox{\textwidth}{!}{
\begin{tabular}{l|ccccccc:c|cccc:c|cccccccc:c|c}
\toprule
& \multicolumn{8}{c|}{Natural}&\multicolumn{5}{c|}{Specialized}&\multicolumn{9}{c}{Structured}&\\
Method& {\rotatebox[origin=l]{90}{CIFAR-100}}&{\rotatebox[origin=l]{90}{Caltech101}}&{\rotatebox[origin=l]{90}{DTD}}&{\rotatebox[origin=l]{90}{Flowers102}}&{\rotatebox[origin=l]{90}{Pets}}&{\rotatebox[origin=l]{90}{SVHN}}&{\rotatebox[origin=l]{90}{Sun397}}&{\rotatebox[origin=l]{90}{Mean}}&{\rotatebox[origin=l]{90}{Camelyon}}&{\rotatebox[origin=l]{90}{EuroSAT}}&{\rotatebox[origin=l]{90}{Resisc45}}&{\rotatebox[origin=l]{90}{Retinopathy}}&{\rotatebox[origin=l]{90}{Mean}}&{\rotatebox[origin=l]{90}{Clevr-Count}}&{\rotatebox[origin=l]{90}{Clevr-Dist}}&{\rotatebox[origin=l]{90}{DMLab}}&{\rotatebox[origin=l]{90}{KITTI-Dist}}&{\rotatebox[origin=l]{90}{dSpr-Loc}}&{\rotatebox[origin=l]{90}{dSpr-Ori}}&{\rotatebox[origin=l]{90}{sNORB-Azim}}&{\rotatebox[origin=l]{90}{sNORB-Elev}}&{\rotatebox[origin=l]{90}{Mean}}&{\rotatebox[origin=l]{90}{Overall Mean}}\\
\multicolumn{24}{c}{CLIP backbone}\\
AdaptFormer     & 73.7 &93.2 &75.2 &96.8 &90.7 &92.7 &56.1 &82.6 &83.3 &95.7 &87.8 &73.6 &85.1 &76.5 &61.9 &49.6 &84.1 &84.6 &55.4 &29.5 &45.7 &60.9 &74.0  \\
AdaptFormer+VQT & 71.3 &95.3 &77.1 &96.2 &90.6 &93.3 &51.2 &82.1 &84.8 &96.4 &88.7 &73.4 &85.8 &75.8 &62.6 &52.4 &83.8 &91.8 &54.6 &33.6 &46.5 &62.6 &74.7  \\
VPT             & 66.3 &90.1 &73.7 &94.7 &90.3 &91.6 &56.0 &80.4 &83.3 &93.4 &87.3 &75.6 &84.9 &41.5 &57.5 &52.3 &80.7 &65.1 &54.3 &27.7 &28.4 &50.9 &68.9  \\
VPT+VQT         & 70.8 &95.1 &72.7 &93.8 &89.8 &93.5 &54.8 &81.5 &85.2 &95.7 &89.7 &74.8 &86.3 &52.5 &62.6 &55.3 &84.1 &77.1 &56.4 &34.6 &35.1 &57.2 &72.3  \\
\multicolumn{24}{c}{MAE backbone}\\
AdaptFormer     & 53.5 &90.1 &60.3 &83.3 &81.4 &83.0 &29.6 &68.7 &83.0 &93.9 &74.4 &73.8 &81.3 &77.8 &60.3 &44.0 &79.5 &75.9 &53.1 &30.3 &45.6 &58.3 &67.0  \\
AdaptFormer+VQT & 56.8 &90.4 &63.7 &86.8 &80.7 &89.7 &29.7 &71.1 &84.5 &95.4 &80.9 &72.5 &83.3 &65.9 &58.5 &46.5 &84.0 &82.2 &53.2 &32.1 &51.1 &59.2 &68.6  \\
VPT             & 45.5 &88.9 &62.2 &75.1 &73.2 &75.2 &24.4 &63.5 &80.1 &94.6 &68.3 &73.6 &79.1 &69.5 &58.2 &39.4 &70.8 &53.6 &51.2 &20.4 &25.5 &48.6 &60.5  \\
VPT+VQT         & 48.9 &90.3 &65.2 &87.4 &81.8 &75.9 &26.0 &67.9 &81.4 &95.1 &80.8 &73.6 &82.7 &63.3 &59.2 &44.4 &80.2 &46.5 &52.7 &22.8 &28.4 &49.7 &63.4  \\
\multicolumn{24}{c}{Supervised ImageNet-21K backbone}\\
AdaptFormer     & 79.9 &89.8 &68.5 &98.0 &88.3 &81.4 &54.8 &80.1 &80.3 &95.4 &81.1 &72.3 &82.3 &71.0 &55.0 &42.3 &68.8 &65.9 &45.1 &24.9 &29.8 &50.3 &68.0  \\
AdaptFormer+VQT & 77.1 &93.7 &68.2 &98.2 &89.8 &84.1 &45.9 &79.6 &82.1 &96.2 &85.6 &73.2 &84.3 &71.4 &54.9 &44.5 &72.3 &76.7 &45.2 &27.6 &31.3 &53.0 &69.4  \\
VPT             & 79.8 &89.9 &67.5 &98.0 &87.0 &79.4 &52.3 &79.1 &83.5 &96.0 &83.7 &75.2 &84.6 &68.1 &60.1 &43.0 &74.8 &74.4 &44.4 &30.0 &40.2 &54.4 &69.9  \\
VPT+VQT         & 76.8 &92.6 &69.2 &98.3 &87.8 &81.6 &46.2 &78.9 &81.3 &96.3 &84.7 &72.4 &83.7 &59.6 &60.3 &43.0 &77.6 &79.3 &46.0 &31.2 &39.5 &54.6 &69.7  \\
\bottomrule
\end{tabular}}
\caption{Test accuracies for AdaptFormer, VPT and their combinations with VQT on the VTAB-1k benchmark on ViT-B/16 pre-trained with CLIP, MAE and Supervised ImageNet-21K. "Mean" denotes the average accuracy for each category and "Overall Mean" shows the average accuracy over 19 tasks.}
\label{table:detailed_acc}
\end{table*}

% \subsection{Accuracy vs. Score Layer Importance}

\section{Additional Discussions} \label{suppl-sec:additional_discussions}

\subsection{More Discussions on Memory Usage}

As mentioned in the last paragraph of \autoref{approach: compare} and as shown in \autoref{experiment: memory}, since VQT keeps all the intermediate features intact and only learns to tune the query tokens and the prediction head, the training process bypasses the expensive back-propagation steps and does not require storing any intermediate gradient results, making it very memory-efficient. As shown in \autoref{fig:vpt}, VPT needs to run back-propagation (red arrow lines) through the huge backbone in order to update the inserted prompts. On the contrary, VQT only needs gradients for the query tokens because all intermediate output features are unchanged, as shown in \autoref{fig:vqt}.

%\subsection{Training Efficiency}
%Another key property of VQT that can not be overlooked is its outstanding training efficiency, which can be analyzed in two aspects. On the one hand, since VQT does not change any intermediate features, we can pre-compute the intermediate features for all the data and store them in the hard disk for later training. Given a standard ViT with 12 Transformer layers and 197 embeddings of size 768 for each layer, all the intermediate features for an image have 12*197*768 floats (7MB), and storing them for a task in VTAB with 1K images only requires 7G in the hard disk. With all the pre-computed intermediate features, we can parallelize the forward and backward passes of the 12 layers at the same time, making the training process 12x faster. On the other hand, since VQT only uses the output of the query tokens for predictions, we just need to pass the features related to the query tokens through the MLP block during the forward pass in each layer, making it additionally faster on top of the cross-layer parallelization mentioned above.  

\subsection{Cost of VQT and AdaptFormer}

In \autoref{experiment: compatible}, to confirm that the improvement mentioned above does not simply come from the increase of tunable parameters, we enlarge AdaptFormer's added modules by increasing the bottleneck dimension $\hat{d}$ from 64 to 128 and 256 to match the tunable parameter number of AdaptFormer when equipped with VQT. Here, we show the detailed parameter calculation in \autoref{fig:adapt_vqt}. The additional parameters for AdaptFormer and VQT can be calculated as  $\hat{d} \times 2 \times D \times M$ and $\underbrace{T \times D \times M}_\text{query tokens} + \underbrace{T \times D \times M \times C}_\text{prediction head}$, respectively where $\hat{d}$ denotes  the bottleneck dimension of AdaptFormer; $D$ is the embedding dimension; $M$ is the number of Transformer layer; T represents the number of VQT's query tokens; $C$ denotes the average number of classes in VTAB, and we round it to 50 for simplicity. The numbers of tunable parameters and percentages of tunable parameters over ViT-B's number of parameters (86M) for AdaptFormer and AdaptFormer+VQT are shown in \autoref{tab:cost}.

\subsection{Training Efficiency}
In this subsection, we point out another \emph{potential} advantage of VQT besides its parameter and memory efficiency --- training efficiency (\ie, the number of floating-point operations and the overall wall-clock time in training). This can be seen from two aspects. 

On the one hand, since VQT does not change the original intermediate features obtained from the pre-trained backbone but only learns to combine them, we can pre-compute them for all the downstream data and store them in the hard drive or even random access memory (RAM)\footnote{We note that these are not the same memory as in memory efficiency. The latter refers to the GPU or CPU memory.} for later training (epochs). As mentioned in~\autoref{suppl_sub_train_detail}, we perform $100$ epochs for learning the query tokens in VQT, in which we indeed only need to compute the intermediate features once in the first epoch, and reuse them in later epochs. Given a standard ViT-B with 12 Transformer layers and 197 embedding tokens of size 768 for each layer, all the intermediate features for an image amount to ``$12\times 197\times 768$'' 32-bit floats (7MB); storing them for a task in VTAB with 1K images only requires 7GB in the hard drive or RAM. With all the pre-computed intermediate features, we can parallelize the forward and backward passes of the 12 layers at the same time, potentially making the training process $12\times$ faster.

On the other hand, since VQT only uses the outputs of the newly introduced query tokens for predictions, during the forward pass within each layer, we just need to pass the MSA features corresponding to these tokens to the MLP block, making it additionally faster on top of the cross-layer parallelization mentioned above.

%Another key property of VQT that can not be overlooked is its outstanding training efficiency, which can be analyzed in two aspects. On the one hand, since VQT does not change any intermediate features, we can pre-compute the intermediate features for all the data and store them in the hard disk for later training. Given a standard ViT with 12 Transformer layers and 197 embeddings of size 768 for each layer, all the intermediate features for an image have 12*197*768 floats (7MB), and storing them for a task in VTAB with 1K images only requires 7G in the hard disk. With all the pre-computed intermediate features, we can parallelize the forward and backward passes of the 12 layers at the same time, making the training process 12x faster. On the other hand, since VQT only uses the output of the query tokens for predictions, we just need to pass the features related to the query tokens through the MLP block during the forward pass in each layer, making it additionally faster on top of the cross-layer parallelization mentioned above.  

\begin{table*}
    \centering
    \begin{tabular}{c|ccc}
        \toprule
        AdaptFormer &$\hat{d}=64$ & $\hat{d}=128$ & $\hat{d}=256$ \\
        \midrule
        Tunable parameters \#&1179648 & 2359296 & 4718592\\
        Tunable parameters \%&1.37\%&	2.74\%&	5.49\%\\
        \midrule
        AdaptFormer+VQT&$\hat{d}=64$ \& &$\hat{d}=64$ \& $T=2$&$\hat{d}=64$ \& $T=4$\\
        \midrule
        Tunable parameters \#&1179648 & 2119680 & 3059712\\
        Tunable parameters \%&1.37\%&2.46\%&	3.56\%\\
        
        \bottomrule
    \end{tabular}
    \caption{Numbers of tunable parameters and percentages of tunable parameters over ViT-B's number of parameters (86M) for AdaptFormer with different bottleneck dimensions and AdaptFormer+VQT with different numbers of query tokens.}
    \vskip-10pt
    \label{tab:cost}
\end{table*}

\clearpage

\end{document}